\documentclass[10pt,twocolumn,letterpaper]{article}

\usepackage{wacv}              %

\usepackage{graphicx}
\usepackage{amsmath}
\usepackage{amssymb}
\usepackage{booktabs}
\usepackage{midfloat}
\usepackage{amsfonts}
\usepackage[table]{xcolor}
\usepackage{verbatim}

\usepackage{float}
\usepackage{tikz}

\usepackage{booktabs}
\usepackage{multirow}
\usepackage{cuted}
\usepackage{adjustbox}
\usepackage{capt-of}
\usepackage{subcaption}
\usepackage{pgf}

\usepackage{caption}
\usepackage{xcolor}
\usepackage{bbm}
\usepackage{colortbl}
\usepackage[table]{xcolor}
\definecolor{Pink}{rgb}{1,0.941,1.}
\definecolor{PositiveGreen}{rgb}{0,0.56,0}

\usepackage[pagebackref,breaklinks,colorlinks,citecolor=blue]{hyperref}

\usepackage[capitalize]{cleveref}
\crefname{section}{Sec.}{Secs.}
\Crefname{section}{Section}{Sections}
\Crefname{table}{Table}{Tables}
\crefname{table}{Tab.}{Tabs.}

\def\wacvPaperID{1749} 
\def\confName{WACV}
\def\confYear{2025}

\newif\ifdraft
\drafttrue

\ifdraft
\newcommand{\KK}[1]{{\color{cyan}{\bf KK: #1}}}

\newcommand{\AL}[1]{{\color{blue}{\bf AF: #1}}}

\newcommand{\MS}[1]{{\color{green}{\bf MS: #1}}}

\else
\newcommand{\KK}[1]{}

\newcommand{\AL}[1]{}

\newcommand{\MS}[1]{}

\fi

\begin{document}
\title{
 NAT: Learning to Attack Neurons for Enhanced Adversarial Transferability
}
\author{Krishna Kanth Nakka\\
VITA Lab, EPFL, Switzerland\\
{\tt\small krishkanth.92@gmail.com}
\and
Alexandre Alahi\\
VITA Lab, EPFL, Switzerland\\
{\tt\small alexandre.alahi@epfl.ch}
}
\maketitle

\begin{strip}
\centering
\includegraphics[width=0.95\textwidth]{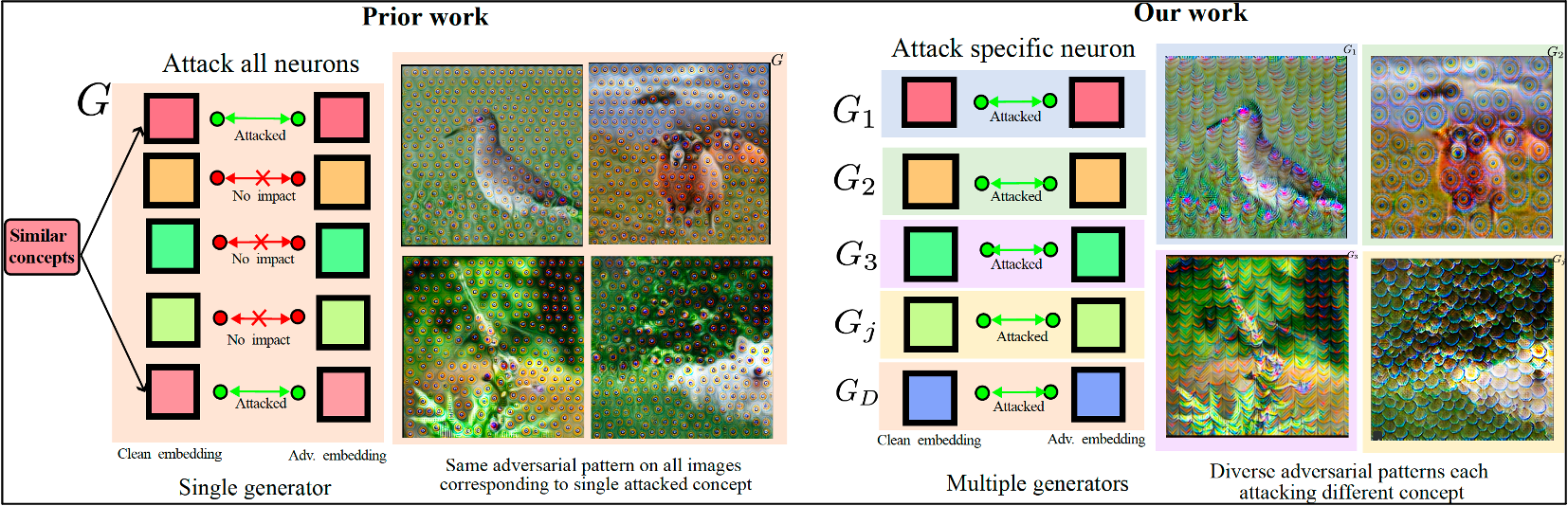}
\captionof{figure}{{\bf Comparison with prior work~\cite{ltp,bia}.} On the left, we illustrate how prior single generator-based methods, such as LTP~\cite{ltp} and BIA~\cite{bia}, attack the entire embedding but predominantly disrupt neurons related to a single concept, like a circular text pattern, while leaving most other neurons largely unaffected. In contrast, our framework trains specific generators to target individual neurons, each representing distinct concepts~\cite{clipdissect}. By focusing on attacking neurons that represent low-level concepts, our method not only generates highly transferable perturbations but also produces diverse, complementary attack patterns. Best viewed in color and zoomed.}
\label{fig:teaser}
\end{strip}

\begin{abstract}
The generation of transferable adversarial perturbations typically involves training a generator to maximize embedding separation between clean and adversarial images at a single mid-layer of a source model. In this work, we build on this approach and introduce Neuron Attack for Transferability (NAT), a method designed to target specific neuron within the embedding. Our approach is motivated by the observation that previous layer-level optimizations often disproportionately focus on a few neurons representing similar concepts, leaving other neurons within the attacked layer minimally affected. NAT shifts the focus from embedding-level separation to a more fundamental, neuron-specific approach. We find that targeting individual neurons effectively disrupts the core units of the neural network, providing a common basis for transferability across different models. Through extensive experiments on 41 diverse ImageNet models and 9 fine-grained models, NAT achieves fooling rates that surpass existing baselines by over 14\% in cross-model and 4\% in cross-domain settings. Furthermore, by leveraging the complementary attacking capabilities of the trained generators, we achieve impressive fooling rates within just 10 queries. Our code is available at: \url{https://krishnakanthnakka.github.io/NAT/}
\end{abstract}

\section{Introduction}\label{sec:Intro}

Transferability of adversarial perturbations~\cite{wang2021admix,rap,attack2fool,wu2018understanding,inputdivattack,nakka2022universal} has received great attention for its ability attack models without white-box access, and particularly concerning for safety~\cite{nakka2020indirect,xie2017adversarial,arnab2018robustness,nakka2020towards} and security-critical applications~\cite{security1,security2,security3}. Numerous approaches have been proposed to enhance transferability, broadly falling into two categories: iterative techniques~\cite{dong2018boosting,wang2021admix,inputdivattack,wang2021feature,wang2021enhancing,huang2019enhancing,nakka2020indirect} to optimize the adversarial images with graident-based stratefies, and  generative attacks~\cite{gap,cda,bia,ltp,facl} leveraging generative models. 
The present work focuses on generative attacks, that has gained prominence in recent years due to its ability to efficiently craft adversarial examples with lower computational overhead compared to iterative methods.
Among generative attacks, LTP~\cite{ltp} and BIA~\cite{bia} focus on attacking mid-level embeddings and have achieved impressive transfer rates across diverse models and domains.  However, these generative attacks often produces a same adversarial pattern irrespective of input images. Upon closer examination of the feature spaces of the generated adversarial images, we identify a key limitation: the generated perturbations disproportionately disrupt only a few neurons\footnote{We follow convention in interpretability literature~\cite{zhang2021survey} and use ``neuron" to describe a single channel in CNN.} representing similar concepts, leaving most neurons within the embedding minimally affected. Consequently, the transferability of these attacks is dependent upon the presence and impact of this single concept in the target model. This observation motivates to investigate the question: \emph{Can an attack that disrupts a broader range of neurons within the embedding result in stronger, more transferable adversarial attacks?}

In response, we introduce a simple yet effective framework called {\bf Neuron Attack for Transferability (NAT)}, which trains a single generator to disrupt specific neuron within the embedding rather than targeting all neurons simultaneously. Our hypothesis builds on the fact that recent works such as CLIP-Dissect~\cite{clipdissect}, Describe-and-Dissect~\cite{dnd} have shown that each neuron within the embedding can represent an interpretable concept (e.g., Neuron 24 in layer 3 of ResNet50 represent vertical stripes), and attempting to disrupt all such concepts with a single adversarial image may be ineffective due to conflicting objectives. In contrast, by focusing on disrupting smaller units—i.e., individual neuron—NAT learns diverse and specialized adversarial perturbation generators. We illustrate this in Figure~\ref{fig:teaser}, where prior methods, BIA and LTP, produce same circular adversarial texture patterns across different images, while NAT generates diverse patterns targeting specific neurons, leading to more effective and varied perturbations.

Furthermore, we go beyond the limited evaluations of transferability on just 5 target models seen in prior works~\cite{ltp,bia,cda} and conduct a rigorous evaluation on 41 ImageNet-pretrained models, covering traditional Convolutional networks~\cite{resnet,squeezenet,mobilenet,mnasnet,xception}, Transformers~\cite{vit,swin,deit,beit}, Hybrid architectures~\cite{convnext,mixer,mobilevit,levit}, and nine fine-grained dataset models~\cite{cub,scars,fgvc}. Through extensive evaluations, we demonstrate that a single neuron-specific adversarial generator achieves over 14\% improvement in transferability for cross-modal settings and 4\% improvement in cross-domain settings. Additionally, by leveraging the complementary attack capabilities of NAT's generators, we show that adversarial transferability can be significantly enhanced with fewer than 10 queries to the target model. \\

\noindent{\bf Contributions.} In summary, we make three key contributions:
\begin{enumerate}
    \item We identify the embedding-level attack primarily focus on disrupting a small set of neurons representing a similar concept, leaving many neurons unaffected inside the embedding.
    \item We introduce a novel strategy of training individual generators for each neuron, significantly enhancing the effectiveness of adversarial perturbations by focusing on a single objective.
    \item We evaluate our NAT framework on over 40 ImageNet-pretrained models, demonstrating that it outperforms baseline methods in standard single-query settings. Additionally, transferability improves by more than 20\% with a small query budget, using diverse adversarial generators.
\end{enumerate}

\begin{table}[t]
    \centering
    \resizebox{\columnwidth}{!}{%
        \begin{tabular}{|c|c|c|c|c|}
            \toprule
            {\bf Method} & Generators & {\bf Attack Layer} & {\bf Neuron} & {\bf Loss Function}  \\
            \midrule
            GAP~\cite{gap} & Single & Final layer & All & $-\texttt{CE}(f_L(\mathbf{x}), y)$  \\
            CDA~\cite{cda} & Single& Final layer &  All & $-\texttt{CE}(f_L(\mathbf{x}) - f_L(\mathbf{x}_a), y)$ \\ \midrule
            LTP~\cite{ltp} & Single& Mid-layer $l$ & All & $-\left\| f_l(\mathbf{x}) - f_l(\mathbf{x}_a) \right\|_2^2$  \\
            BIA~\cite{bia} & Single& Mid-layer $l$ & All & $\frac{f_l(\mathbf{x}) \cdot f_l(\mathbf{x}_a)}{\|f_l(\mathbf{x})\| \|f_l(\mathbf{x}_a)\|}$  \\ \midrule
            NAT (Ours) & Multiple & Mid-layer $l$ & Single $j$ & $- \left\| f_l^j(\mathbf{x}) - f_l^j(\mathbf{x}_a) \right\|_2^2$  \\
            \bottomrule
        \end{tabular}
    }
   \caption{%
   We compare various adversarial attack methods based on the attack layer, the number of neurons targeted, and the loss function used. Our method specifically targets individual neurons $j$ within an intermediate layer $l$ and employs the $L_2$ loss function, similar to LTP~\cite{ltp}.$f_l$ represents the embedding at layer $l$, $\mathbf{x}$ and $\mathbf{x}_a$ represent the clean and adversarial images, $L$ represents the total number of layers, and $y$ represents the ground-truth label.}
    \label{tab:attack_comparison}
\end{table}

\section{Related Work}\label{sec:relatedwork}

\noindent{\bf Adversarial Attacks.}
Deep neural networks are highly vulnerable to adversarial perturbations, where subtle changes to input data cause the model to produce incorrect outputs, as first demonstrated in~\cite{goodfellow2014explaining}. This vulnerability has led to the development of various iterative methods such as BIM~\cite{bim}, PGD~\cite{pgd}, CW~\cite{carlini2019evaluating}, and DeepFool~\cite{moosavi2016deepfool}, which iteratively refine perturbations to deceive target models. Following this, several works~\cite{wang2021enhancing,wang2021feature,wu2021improving} identified that adversarial perturbations could be crafted using substitute models to enable transfer attacks on unknown models. Techniques such as momentum boosting~\cite{dong2018boosting}, feature separation~\cite{yan2022ila,huang2019enhancing,li2020yet}, data augmentation~\cite{wu2021improving,yuan2022adaptive}, and advanced optimization strategies~\cite{wang2021enhancing,rap} were introduced to improve transferability. However, these methods rely on computationally expensive iterative optimization and have shown limited transferability when attacking target models with significant architectural differences from the source model.

\noindent{\bf Generative Attacks.}
Generator-based attacks offer an alternative approach, using generative models to craft adversarial examples~\cite{gap,cda,ltp,bia} by leveraging training data. Early work in this area, such as GAP~\cite{gap}, trained a generator to maximize cross-entropy loss against the ground truth labels. This was later refined by methods like CDA~\cite{cda}, which introduced a relativistic cross-entropy loss. Further advancements such as LTP~\cite{ltp} and BIA~\cite{bia} improved these approaches by focusing on maximizing embedding-separation loss at specific intermediate layers of the source model during generator training. Recent approaches like GAMA~\cite{gama} and PDCL~\cite{pdcl} leveraged the powerful text supervision from CLIP~\cite{clip} to guide the generator, providing additional training signals alongside the source model. In other direction, ~\cite{facl} proposed FACL by incorporating frequency-aware feature separation along with standard training. A comparative analysis of key representative methods is summarized in Table~\ref{tab:attack_comparison}. Notably, most of these works~\cite{ltp,bia,facl,gama,pdcl} share a underlying common theme of using of embedding separation loss between the original and adversarial embeddings of a specific intermediate layer to guide the generator training process.

\noindent{\bf Our Work.}
In contrast to prior methods, our approach shifts the focus of perturbation learning from a embedding-level strategy~\cite{ltp,bia} to a more granular, neuron-specific one~\cite{dnd,clipdissect}. We demonstrate that generators targeting individual neurons are not only more effective due to their focused objective but also complement each other effectively, leading to a significant improvement in empirical performance, both in standard single-query settings and in boosting transfer rates within just a few queries.

\begin{figure}[t!]
\centering
\includegraphics[width=\columnwidth]{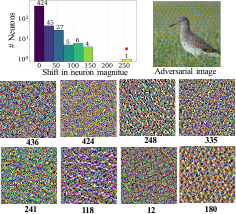}
\caption{In the top row, we present the histogram of activation shifts between clean and adversarial images for each neuron in the attacked layer $l=18$ using LTP~\cite{ltp}, along with a corresponding sample adversarial image. The bottom two rows display the synthesized images of the top-8 most disrupted neurons, which are closely associated with the circular patterns observed in the adversarial image (top right). The numbers below the images in the bottom rows indicate the neuron positions within the 512-D embedding. Best viewed in color and with zoom.}\label{fig:revisit_attacks} \label{fig:revisit_attacks}
\end{figure}

\section{Method} \label{sec:Method}

We propose the NAT framework to create transferable adversarial perturbations aimed at disrupting individual neurons within a specific intermediate layer. Before detailing our method, we revisit existing feature-based attacks~\cite{ltp,bia} and analyze the feature spaces of the generated adversarial images to motivate our approach.

\subsection{\bf Revisiting Feature-Based Attacks}

Let $\mathbf{x} \in \mathbb{R}^{H \times W \times 3}$ represent the input image. A generator $\mathbf{G}_{\theta}(\cdot)$ perturbs this image to produce $\mathbf{x}_a \in \mathbb{R}^{H \times W \times 3}$ within the perturbation budget $\epsilon$. The surrogate model $f$ generates embeddings $f_l(\mathbf{x})$ and $f_l(\mathbf{x}_a)$ at layer $l$, with dimensions $\mathbb{R}^{H_l \times W_l \times D_l}$, where $D_l$ represents the number of neurons (channels). The generator in LTP~\cite{ltp} is trained to maximize the separation between these embeddings using the $L_2$ loss function, $\mathcal{L}_{{adv}}(G) = \left\| f_l(\mathbf{x}) - f_l(\mathbf{x}_a) \right\|_2^2$. The choice of layer $l$ is critical for transferability and its position is dependent of the source model.

A striking observation from generative attacks~\cite{ltp,bia} is that they consistently produce adversarial images with nearly identical patterns, regardless of the input image (see left side of Figure ~\ref{fig:teaser}). The primary reason for this phenemenon is because the generator disproportionately disrupts only a few neurons that represent similar concepts, leaving most neurons unaffected. We illustrate this in the top row of Figure~\ref{fig:revisit_attacks} with a histogram of the $L_1$ distance between the original and adversarial embeddings, showing that only a single neuron (highlighted by the yellow bar on the right) is significantly disrupted, with a handful of others showing moderate disruption and around 424/512 (83\%) neurons are largely unaffected with the LTA. In the bottom two rows, we visualize the synthesised input images that activate the top-8 most disrupted neurons, obtained using the activation maximization algorithm~\cite{actimax}. We find that these neurons share similar synthethesised images, indicating that they are connected to the same underlying concept, of circular texture patterns. Furthermore, the visual correlation between these synthesized images to the adversarial patterns in the generated images reinforces the observation that generator $G$ predominately disrupts small set of neurons representing a single, closely related concept. This limitation highlights the need for a more comprehensive strategy, which our method addresses by targeting a broader range of neurons independently to enhance transferability.

\begin{figure*}[t!]
\centering
\includegraphics[width=0.9\linewidth]{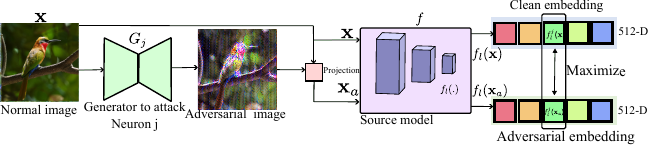}
\caption{Overview of the NAT method. A neuron-specific generator $G_j$ is trained to attack neuron $j$ in layer $l$ of the source model $f$.}    
\label{fig:overview} 
\end{figure*}

\subsection{\bf Neuron-Specific Adversarial Generators}

The observations above reveal a limitation of current feature-based generative attacks: they focus on few and similar neurons, limiting the diversity and effectiveness of adversarial perturbations. To address this, we introduce a simple yet effective strategy that trains individual generators to target specific neuron within the feature map, where each neuron corresponds to a distinct concept. By attacking individual neurons, our approach improves transferability to unseen models while increasing the diversity of generated perturbations. Below, we outline the architecture and training methodology of our framework.

\noindent{\bf Overview.} Figure~\ref{fig:overview} illustrates our framework, which modifies the generator training process to optimize for specific neuron. For each neuron position $j$, we first extract the feature $f_l^j(\mathbf{x}) \in \mathbb{R}^{H_l \times W_l}$ at position $j$ in the embedding $f_l$ of the original image $\mathbf{x}$. We pass the input image to the generator $G_j$ and obtain adversarial image $\mathbf{x}_a = G_j(\mathbf{x})$ whose feature $f_l^j(\mathbf{x}_a)$ at position $j$ is used for training. The generator is trained by maximizing the separation between $f_l^j(\mathbf{x})$ and $f_l^j(\mathbf{x}_a)$. The loss function guiding this process is:

\begin{equation}
    \mathcal{L}_{{adv}}(G^j) = \left\| f_l^j(\mathbf{x}) - f_l^j(\mathbf{x}_a) \right\|_2^2
\end{equation}

Our neuron-specific approach ensures that each generator disrupts a distinct concept~\cite{clipdissect,dnd} represented by a neuron $j$. Since our method focuses on disrupting a single neuron at a time, it operates as a specialized expert, with a specific objective rather than on a broad target to disrupt all concepts. Our approach adapts the ``mixture of experts" structure~\cite{moe2,moe1}, where each generator serves as an expert in attacking a particular concept within the neural network. During inference, the test image can be passed through the trained generators, $G = \{G_1, G_2, \dots, G_{D_l}\}$, producing a set of adversarial images, each target a specific concept. 

\subsection{\bf Implementation Details}\label{sec:implementation}
We use VGG16~\cite{vgg} as the source model $f$ and select the same layer position $l=18$, as recommended by LTP~\cite{ltp}, which consists of $D_l=512$ neurons, for optimal transferability. This allows us to train $D_l$ neuron-specific generators, each targeting a distinct neuron within the embedding $f_l$. However, due to the substantial computational cost of training 512 generators on the full dataset, we adopt a low-fidelity evaluation strategy, commonly used in hyperparameter optimization techniques~\cite{li2018hyperband,yang2020hyperparameter}. Initially, all 512 generators are trained on a small subset (3.12\%) of the ImageNet training set, requiring just 15 minutes for each generator's training. After this initial lightweight training phase, we select the top $k=40$ generators that exhibit the highest transferability, by evaluating on the same source model ie., VGG16~\cite{vgg} or by using a single held-out model. The selected top $k=40$ perturbation generators are then fully trained on the complete ImageNet training set. It is worth noting that even baseline methods such as LTP~\cite{ltp} and BIA~\cite{bia} require at least one held-out model to determine the optimal layer $l$ position.

\section{Experiments}\label{sec:experiments}

We first introduce the datasets and target models used in the experiments, followed by key results in both cross-model and cross-domain settings. Additionally, we provide qualitative analyses to explore the underlying reasons for transferability, and conclude with deeper insights into the performance of the trained generators.

\noindent{\bf Datasets and Models.} 
We follow similar experimental setting as in prevous works~\cite{cda,bia,ltp}. We use the ImageNet training set~\cite{imagenet} for training the generatros. Similar to prior works~\cite{bia,cda,ltp}, we use VGG16~\cite{vgg} as the source model and select the 18th layer as the attack layer $l$, following the recommendation in LTP~\cite{ltp}. We evaluate our method in two settings: 1. cross-model and 2. cross-domain. For the cross-model setting, we use a diverse pool of 41 ImageNet-trained architectures, including 16 convolutional networks~\cite{resnet,densenet}, 7 vision transformer-based models~\cite{vit,deit}, and 18 hybrid/efficient models~\cite{mixer,conformer,convnext}. For evaluation dataset, we use the same subset of 5K samples from the ImageNet validation set released by LTP~\cite{ltp}. Next, for the cross-domain setting, we test transferability across three fine-grained datasets: CUB200~\cite{cub}, FGVC~\cite{fgvc}, and Stanford Cars~\cite{scars}, following the methodology in BIA~\cite{bia}. We utilize pretrained models from the TorchVision~\cite{torchvision2016} and TiMM libraries~\cite{rw2019timm}, with specific model versions provided in the supplementary. \\

\noindent{\bf Baselines.} 
We compare our approach with four state-of-the-art methods: GAP~\cite{gap}, CDA~\cite{cda}, LTP~\cite{ltp}, and BIA~\cite{bia}. GAP~\cite{gap} minimizes cross-entropy loss, while CDA~\cite{cda} uses relativistic cross-entropy loss. LTP~\cite{ltp} and BIA~\cite{bia} focus on attacking mid-level features, using $L_2$ loss and cosine loss, respectively. We use the pretrained generators provided by the authors, except for GAP, for which we trained the models due to the lack of public models. The perturbation generator is trained with a learning rate of $0.0002$, using beta values of $0.5$ and $0.999$ for the Adam~\cite{adam} optimizer with batch size set to 16. To ensure reproducibility, we replace non-deterministic \texttt{ReflectionPadding} layers in the generator with the padding operation. We set the perturbation budget $\epsilon$ to 10~\cite{ltp,cda,bia}. \\

\noindent{\bf Neuron Selection.} In the main experiments, we assume access to DenseNet121~\cite{densenet} to select the top-$k=40$ neurons out of 512 neurons after lightweight training with a small subset (3.12\%) of training data explained in \textsection~\ref{sec:implementation}. Additionally, we conduct an ablation study (\textsection~\ref{sec:discussion}) on neuron selection using different heldout models, including the source model, showing that even without access to different held-out model, we achieve superior performance. Nevertheless, we note that even prior feature-based attack, LTP~\cite{ltp} and BIA~\cite{bia} require held-out models to tune the optimal layer $l$ at least once. We provide the exact neuron positions $j$ and their ranks, used to train our top-$k$ generators across different held-out models, in the supplementary.\\

\noindent{\bf Metrics.} Following ~\cite{ltp,cda,bia}, we report the fooling rate, defined as the percentage of images for which the predicted label is modified. In cross-domain experiments, we report the adversarial accuracy metric, in line with BIA~\cite{bia}. Importantly, we attack the target model using the top-$k=40$ trained generators, in the same order as ranked by the held-out model to select top-$k$ neurons (\textsection~\ref{sec:implementation}). When attacking with $k$-queries, we assume the attack as the successful if atleast one of the query is successful. %

\begin{figure}[t]
\centering
\includegraphics[width=\columnwidth]{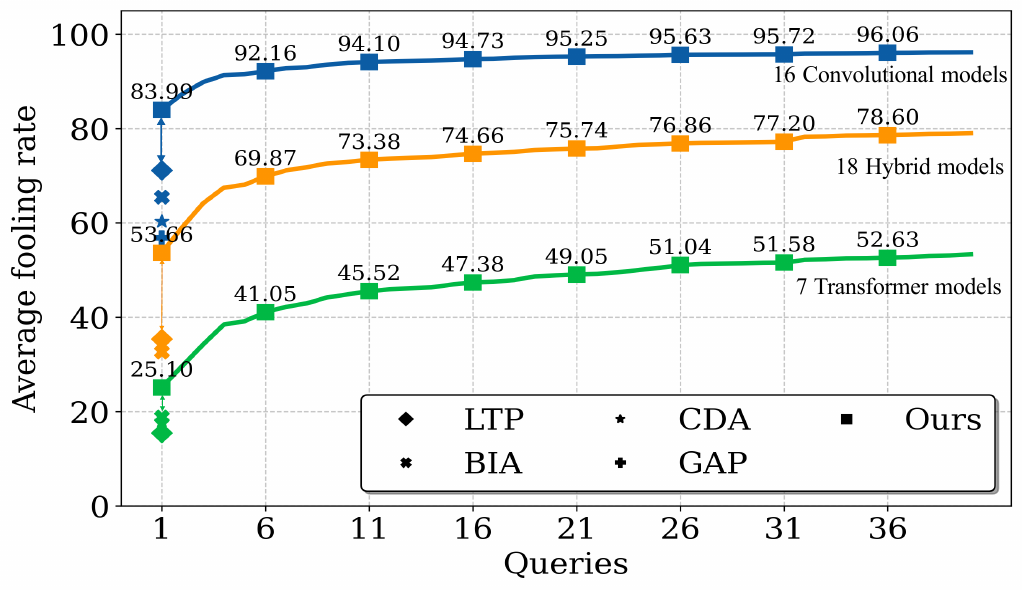}
\caption{{\bf Performance of NAT.} We report the average fooling rates for three groups of models: 16 Convolutional models, 7 Transformer models, and 18 Hybrid models, represented by blue, orange, and green lines, respectively. Our method outperforms the baselines in the single-query ($k=1$) setting and achieves a significant boost in transfer rates within just 10 queries, with the performance saturating after 20 queries.}
\label{fig:transfer_rate_vs_query}
\end{figure}

\subsection{Cross-Model Results}

In all prior works~\cite{cda,ltp,bia}, the evaluation of transferability rates has been limited to fewer than five target models. We significantly extend this evaluation setting to 41 ImageNet-pretrained models, allowing for a more comprehensive understanding of attack efficacy across a diverse range of architectures. To facilitate this analysis, we make a reasonable classification of the target models into three categories: convolutional architectures~\cite{resnet,densenet,shufflenet,squeezenet,mobilenet} (16 models), transformer architectures~\cite{swin,deit,beit} (7 models), and hybrid architectures~\cite{cait,mixer,focalnet} (18 models).\\

The core findings of this paper are presented in Figure~\ref{fig:transfer_rate_vs_query}, where we compare our attack performance with baselines in both single-query settings and across increasing query budgets using 40 generators ($k=40$ neurons) for the model groups shown in blue, orange and yellow for Convolution, Transformer and Hybrid architectures, respectively. The results clearly indicate that our method not only surpasses baseline approaches under a one-query budget but also demonstrates substantial gains in effectiveness with a limited number of additional queries. A detailed discussion of the outcomes for each model group is provided below.\\

\noindent {\bf Convolutional Models.} We report the fooling rates for the 16 convolutional models in Table~\ref{tab:conv_results}. Our NAT method outperforms baselines by a large margin. Specifically, under a single-query budget, our method achieves an average transfer rate of 84.0\% across 16 architectures, significantly higher than the transfer rates achieved by GAP (59.2\%), BIA (64.8\%), CDA (60.9\%), and LTP (71.5\%). Furthermore, our method gives impressive results with small query budget, achieving {\bf 93.9\%} and {\bf 96.2\%} fooling rates with 10 and 40 queries, respectively, highlighting the diversity of our trained generators. This shows that although models different in terms of base architecture, the underlying neuron representations within the models are similar, thus faciliating the transfer with just 10 queries.\\

\noindent {\bf Transformer Models.} The results for the 7 transformer models, presented in Table~\ref{tab:transformer_results}, further corroborate the efficacy of our attack. Despite the architectural differences between self-attention mechanisms and convolutional operation, our method achieves an average transfer rate of {\bf 25.0\%} across 7 transformer models in a single-query setting, outperforming the best baseline BIA of 18.9\%. With 10 and 40 queries, the fooling rates increase to 44.9\% and 53.5\%, respectively, showcasing the effective transferability from VGG16 convolutional backbone to diverse transformers such as ViT, DeiT, and BeiT.\\

\noindent {\bf Hybrid Models.} For the 18 models with hybrid convolutional-transformer architectures, our method continues to demonstrate superiority. As detailed in Table~\ref{tab:mixed_results}, our method achieves an average transfer rate of {\bf 53.6\%} with one query, surpassing LTP by 18.2 percentage points. Following the similar trend, the fooling rates increase to 74.1\% with 10 queries and 79.1\% with 40 queries, further highlighting the strength of our attack on hybrid architectures.

\begin{table}[ht]
\centering
\small
\resizebox{0.9\columnwidth}{!}{%
\begin{tabular}{l|c|c|c|c|c}
\toprule
\textbf{Model} & \textbf{GAP $\uparrow$} & \textbf{CDA $\uparrow$} & \textbf{BIA $\uparrow$} & \textbf{LTP $\uparrow$} & \textbf{Ours (1/10/40) $\uparrow$} \\
\midrule
	ResNet152~\cite{resnet} & 49.6 & 52.2 & 55.7 & 68.5 & \textbf{89.1} / 97.1 / 97.8 \\ 
DenseNet121~\cite{densenet} & 80.9 & 62.7 & 73.9 & 84.6 & \textbf{91.0} / 98.5 / 98.8 \\ 
SqueezeNet1~\cite{squeezenet} & 66.1 & 70.7 & 88.6 & 86.0 & \textbf{89.5} / 97.6 / 99.1 \\ 
ShuffleNet~\cite{shufflenet} & 50.7 & 50.5 & 68.5 & 67.7 & \textbf{89.0} / 97.4 / 98.2 \\ 
MNasNet~\cite{mnasnet} & 75.3 & 84.9 & 87.4 & 90.6 & \textbf{93.3} / 98.8 / 99.5 \\ 
PNasNet~\cite{pnasnet} & 57.3 & 37.2 & 45.9 & 33.8 & \textbf{63.6} / 79.8 / 86.9 \\ 
MobileNet~\cite{mobilenet} & 51.6 & 63.1 & 63.3 & 60.2 & \textbf{84.3} / 95.1 / 97.0 \\ 
WideResNet~\cite{wideresnet} & 65.7 & 65.0 & 67.5 & 84.9 & \textbf{92.2} / 98.3 / 98.8 \\ 
RegNet~\cite{regnet} & 61.3 & 79.8 & 63.8 & 82.1 & \textbf{91.2} / 98.5 / 98.7 \\ 
AlexNet~\cite{alexnet} & 50.7 & 48.2 & 50.9 & 42.2 & \textbf{49.8} / 70.7 / 83.1 \\ 
GhostNetV2~\cite{ghostnet} & 69.5 & 63.1 & 63.9 & 73.4 & \textbf{82.7} / 96.4 / 98.4 \\ 
CSPResNet~\cite{cspresnet} & 42.4 & 42.8 & 46.8 & 72.8 & \textbf{83.4} / 94.2 / 95.6 \\ 
HRNet~\cite{hrnet} & 66.1 & 83.7 & 83.0 & 93.8 & \textbf{97.1} / 99.3 / 99.8 \\ 
Xception~\cite{xception} & 45.9 & 40.7 & 56.8 & 55.8 & \textbf{73.8} / 88.6 / 93.0 \\ 
NF ResNet~\cite{nfresnet} & 34.5 & 41.2 & 38.5 & 52.6 & \textbf{76.9} / 92.5 / 94.4 \\ 
DLA & 80.2 & 88.6 & 82.6 & 94.3 & \textbf{97.9} / 99.6 / 99.8 \\ 
\midrule
Average & 59.2 & 60.9 & 64.8 & 71.5 & \textbf{84.0} / 93.9 / 96.2   \\ 
\bottomrule
\end{tabular}}
\caption{{\bf Transferability on Convolutional Models.}}\label{tab:conv_results}
\end{table}

\begin{table}[ht]
\centering
\small
\resizebox{0.9\columnwidth}{!}{%
\begin{tabular}{l|c|c|c|c|c}
\toprule
\textbf{Model} & \textbf{GAP $\uparrow$} & \textbf{CDA $\uparrow$} & \textbf{BIA $\uparrow$} & \textbf{LTP $\uparrow$} & \textbf{Ours (1/10/40) $\uparrow$} \\
\midrule
ViT\_B\_16~\cite{vit} & 17.5 & 16.8 & 19.2 & 11.7 & \textbf{18.0} / 33.2 / 43.8 \\ 
ViT\_L\_16~\cite{vit} & 18.0 & 18.1 & 18.8 & 13.4 & \textbf{18.1} / 34.4 / 44.9 \\ 
Swin~\cite{swin} & 14.7 & 18.9 & 21.4 & 17.9 & \textbf{28.8} / 53.6 / 62.4 \\ 
BeIT~\cite{beit} & 16.1 & 17.1 & 15.3 & 12.4 & \textbf{30.5} / 46.6 / 55.2 \\ 
DeiT~\cite{deit} & 13.5 & 15.0 & 16.8 & 12.3 & \textbf{20.5} / 38.7 / 46.7 \\ 
Cait~\cite{cait} & 16.5 & 17.4 & 18.4 & 14.7 & \textbf{27.6} / 46.4 / 54.7 \\ 
Davit~\cite{davit} & 15.4 & 22.4 & 22.3 & 25.8 & \textbf{31.6} / 61.5 / 66.8 \\ 
\midrule
Average & 16.0 & 17.9 & 18.9 & 15.5 & \textbf{25.0} / 44.9 / 53.5   \\ 

\bottomrule
\end{tabular}}
\caption{{\bf Transferability on Transformer models}}\label{tab:transformer_results}
\end{table}

\begin{table}[ht]
\centering
\small
\resizebox{0.9\columnwidth}{!}{%
\begin{tabular}{l|c|c|c|c|c}
\toprule
\textbf{Model} & \textbf{GAP $\uparrow$} & \textbf{CDA $\uparrow$} & \textbf{BIA $\uparrow$} & \textbf{LTP $\uparrow$} & \textbf{Ours (1/10/40) $\uparrow$} \\
\midrule
ConvNext~\cite{convnext} & 44.3 & 48.4 & 51.3 & 46.1 & \textbf{70.9} / 93.2 / 97.0 \\ 
Mixer~\cite{mixer} & 24.6 & 27.8 & 31.7 & 26.3 & \textbf{32.4} / 58.7 / 66.1 \\ 
ConvMixer~\cite{convmixer} & 35.8 & 48.6 & 36.3 & 46.0 & \textbf{76.1} / 91.4 / 94.5 \\ 
CrossViT~\cite{crossvit} & 14.5 & 15.9 & 16.9 & 12.8 & \textbf{19.2} / 34.8 / 44.1 \\ 
Edgenext~\cite{edgenext} & 33.0 & 47.2 & 44.4 & 47.9 & \textbf{73.3} / 88.7 / 91.0 \\ 
GCViT~\cite{gcvit} & 16.1 & 22.6 & 23.5 & 31.0 & \textbf{44.5} / 66.2 / 72.3 \\ 
Visformer~\cite{visformer} & 28.7 & 44.0 & 31.8 & 44.6 & \textbf{71.1} / 91.3 / 93.1 \\ 
FocalNet~\cite{focalnet} & 19.2 & 25.8 & 27.8 & 33.4 & \textbf{45.7} / 72.2 / 82.1 \\ 
Hiera~\cite{hiera} & 20.0 & 24.6 & 22.3 & 21.0 & \textbf{43.2} / 64.5 / 70.3 \\ 
MaxVit~\cite{maxvit} & 12.7 & 16.9 & 18.4 & 17.7 & \textbf{22.9} / 39.8 / 46.7 \\ 
Conformer~\cite{conformer} & 24.0 & 36.9 & 25.7 & 29.6 & \textbf{47.7} / 73.3 / 80.5 \\ 
MobileViT~\cite{mobilevit} & 62.3 & 71.0 & 70.3 & 81.8 & \textbf{92.1} / 98.3 / 99.1 \\ 
RDNet~\cite{rdnet} & 18.0 & 26.5 & 24.9 & 26.7 & \textbf{50.7} / 71.4 / 75.5 \\ 
LeViT~\cite{levit} & 29.0 & 34.0 & 33.2 & 39.5 & \textbf{67.5} / 87.9 / 89.9 \\ 
MViT~\cite{mvit} & 16.8 & 22.0 & 21.5 & 20.2 & \textbf{35.4} / 60.7 / 67.3 \\ 
Coat~\cite{coat} & 22.8 & 31.5 & 30.3 & 33.2 & \textbf{52.1} / 78.7 / 81.3 \\ 
EfficientViT~\cite{efficientvit} & 39.5 & 42.4 & 42.1 & 41.4 & \textbf{60.0} / 83.1 / 88.5 \\ 
EfficientNet~\cite{efficientnet} & 28.7 & 30.2 & 36.4 & 37.3 & \textbf{60.6} / 79.8 / 84.8 \\ 
\midrule
Average & 27.2 & 34.2 & 32.7 & 35.4 & \textbf{53.6} / 74.1 / 79.1   \\ 

\bottomrule
\end{tabular}}
\caption{{\bf Transferability on Hybrid or Efficient Models.}}\label{tab:mixed_results}
\end{table}

\begin{table*}[!t]
\fontsize{10pt}{10pt}\selectfont
\setlength{\tabcolsep}{1.5pt}
\centering
\resizebox{0.8\linewidth}{!}{
\begin{tabular}{ccccccccccccc}
\toprule
\multirow{2}{*}{\textbf{Method}} & \multicolumn{3}{c}{\textbf{CUB-200-2011}~\cite{cub}} && \multicolumn{3}{c}{\textbf{FGVC Aircraft~\cite{fgvc}}} && \multicolumn{3}{c}{\textbf{Stanford Cars}~\cite{scars}} & \multirow{2}{*}{\textbf{AVG.}} \\
\cmidrule{2-12}
& \textbf{Res-50} & \textbf{SENet154} & \textbf{SE-Res101} && \textbf{Res-50} & \textbf{SENet154} & \textbf{SE-Res101} &&  \textbf{Res-50} & \textbf{SENet154} & \textbf{SE-Res101} \\
\midrule
Clean & 87.33 & 86.81 & 86.57 && 92.14 & 92.05 & 91.81 && 94.25 & 93.36 & 92.97 & 90.81 \\ \midrule
GAP~\cite{gap} & 71.23 & 66.95 & 67.43 && 70.93 & 62.17 & 60.55 && 69.15 & 75.75 & 83.85 & 69.78 \\
CDA~\cite{cda} & 71.63 & 64.98 & 73.70 && 73.54 & 58.93 & 71.08 && 70.64 & 68.40 & 83.09 & 70.67 \\
BIA~\cite{bia} & 32.72 & 53.04 & 57.96 && 28.92 & 60.49 & 46.59 && 39.31 & 69.72 & 69.66 & 50.93 \\
LTP~\cite{ltp} & 23.56 & 52.57 & 61.44 && 20.13 & 61.27 & 60.97 && 28.60 & 65.99 & 76.02 & 50.06 \\
FACL~\cite{facl} & {24.74} & {44.06} & {53.75} && {26.58} & {65.71} & {61.40} && {19.72} & {52.01} & {48.51} & {44.05} \\     \midrule
{\bf Ours (k=1)} & 36.52 & 37.64 & 49.65 && 30.60 & 35.22 & 45.39 && 23.58 & 49.30 & 60.59 & {\bf 40.94} \\
{\bf Ours (k=10)} & 11.58 & 17.14 & 31.38 && 3.51 & 7.50 & 26.37 && 5.05 & 19.08 & 40.09 & 17.97 \\
{\bf Ours (k=40)} & 5.97 & 12.06 & 24.78 && 0.54 & 3.84 & 18.06 && 2.77 & 15.04 & 28.32 & 12.38 \\
\bottomrule
\end{tabular}
}
\caption{{\bf Evaluation in Cross-Domain Setting}. We report the adversarial accuracy (in \%) across three fine-grained datasets. We observe that our generators substantially outperform the baseline models in deceiving the target networks in single query $k=1$ and multi-query $k=10$ and $k=40$ settings. }
\label{tab:cross_domain}
\end{table*}

\subsection{Cross-Domain Results}
Similar to previous work~\cite{ltp}, we conduct experiments in the cross-domain setting, where the target models are trained on three fine-grained datasets using three different backbones. Note that the dataset used to train the target models in this setting differs from the one used to train our perturbation generators. The results are summarized in Table~\ref{tab:cross_domain}. Under a single-query budget ($k=1$), our NAT method demonstrates superior performance, achieving an average adversarial accuracy of 40.94\%, compared to 69.8\%, 70.7\%, 50.9\%, and 50.0\% for GAP, CDA, BIA, and LTP, respectively. Moreover, we outperform the recently proposed FACL method~\cite{facl}, which applies frequency-domain supervision, by an additional 3.5\% points. Furthermore, our method achieves an average adversarial accuracy of {\bf 17.97\%} and {\bf 12.38\%} with query budgets of 10 and 40, respectively. These results indicate that our perturbation generators are both diverse and highly effective for disrupting target model predictions even in cross-domain setting.

\subsection{Qualitative Results}

\begin{figure*}[ht!]
\centering
\includegraphics[width=\textwidth]{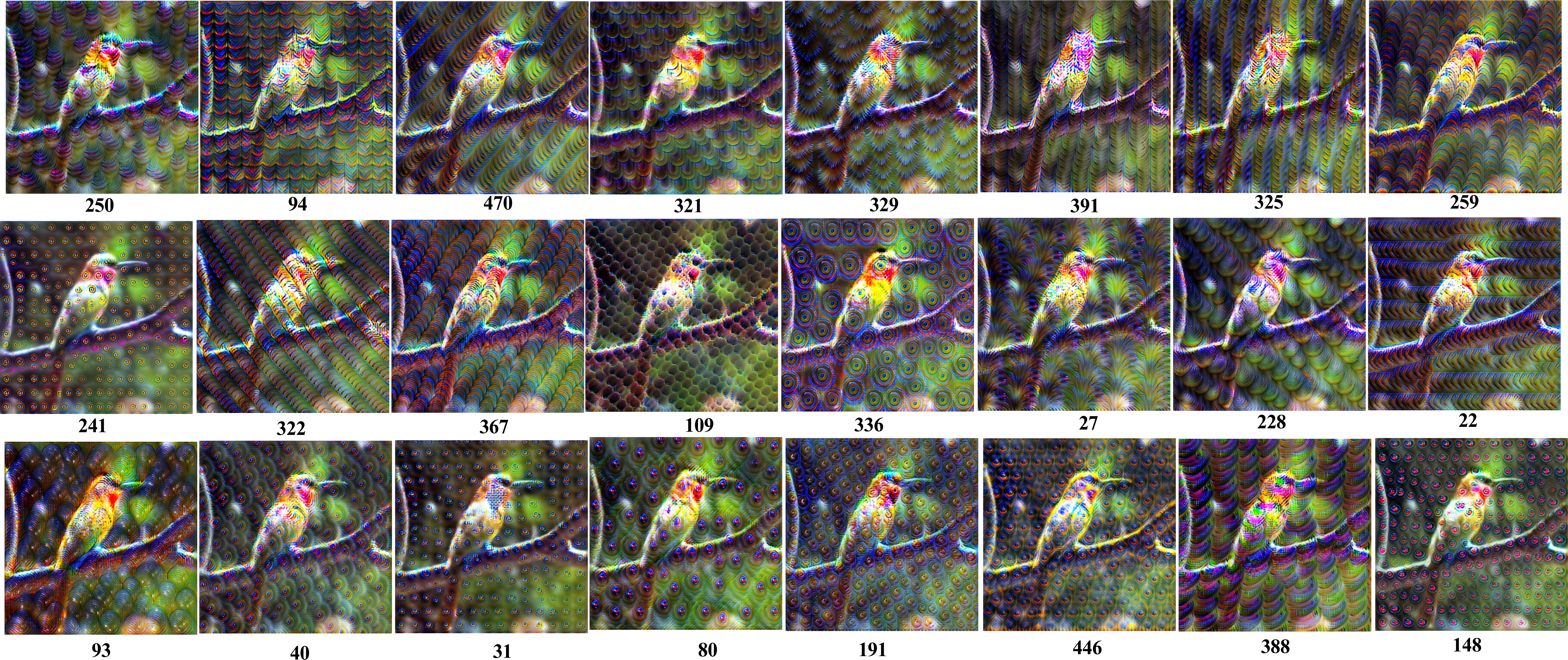}
\caption{{\bf Unbounded Adversarial Images with Our NAT Framework.} We visualize the generated adversarial images before projection using our top-24 neuron-specific generators. Each generator produces a distinct perturbation with its own texture type, complementing one another to achieve superior attack efficacy. The attacked neuron position $j$ associated with each generator is displayed next to each image. We use DenseNet121~\cite{densenet} as the heldout model to choose top-$k$  neuron locations after lightweight training. Best viewed in color and zoomed.}
\label{fig:qualitative}
\end{figure*}

Figure~\ref{fig:qualitative} presents the visualization of unbounded adversarial images from the top-$k$ generators using our NAT framework with the DenseNet121 as the heldout model for neuron selection after lightweight training. Each generated adversarial image demonstrates a unique pattern, corresponding to the neuron attacked by its generator. Furthermore, the diversity in the perturbations allows the attacker to leverage complementary attack capabilities, significantly enhancing the effectiveness of the attack within a limited query budget as observed in cross-domain and cross-model transferability results. Furthermore, LTP and BIA produce pattern that resemble to the ones produced by the generator attacking the position 241 (see first image in second row of Figure~\ref{fig:qualitative}), whereas with our method, we produce diverse ones including the one generated in the prior works.

\subsection{Discussion} \label{sec:discussion}

\noindent {\bf Choosing a Different Held-Out Model.} In all the previous results, we selected the top-$k=40$ neurons based on their transferability to a held-out DenseNet121~\cite{densenet} model after training generators on just 3.12\% of the training dataset. To assess the robustness of this choice of held-out model, we varied it to a different architecture, ResNet152~\cite{resnet}, and also evaluated the worst-case scenario by using the VGG16~\cite{vgg} source model to select the top-$k$ neuron locations. The results of this ablation study are summarized in Table~\ref{tab:ablate_held_out}, with $k=40$ queries in both cross-model and cross-domain settings. 

We observed that in the cross-model setting, the performance using ResNet152 and VGG16 remained within a 2\% range on average across 41 ImageNet models. However, in the cross-domain setting, using ResNet152 as the held-out model yielded better performance of 9.5\%  in average adversarial accuracy over 9 models. However, it is important to note that the cross-domain setting includes only three models per dataset all of which are derived from the ResNet backbone, potentially introducing bias.

\begin{figure}[h!]
\centering
\includegraphics[width=\columnwidth]{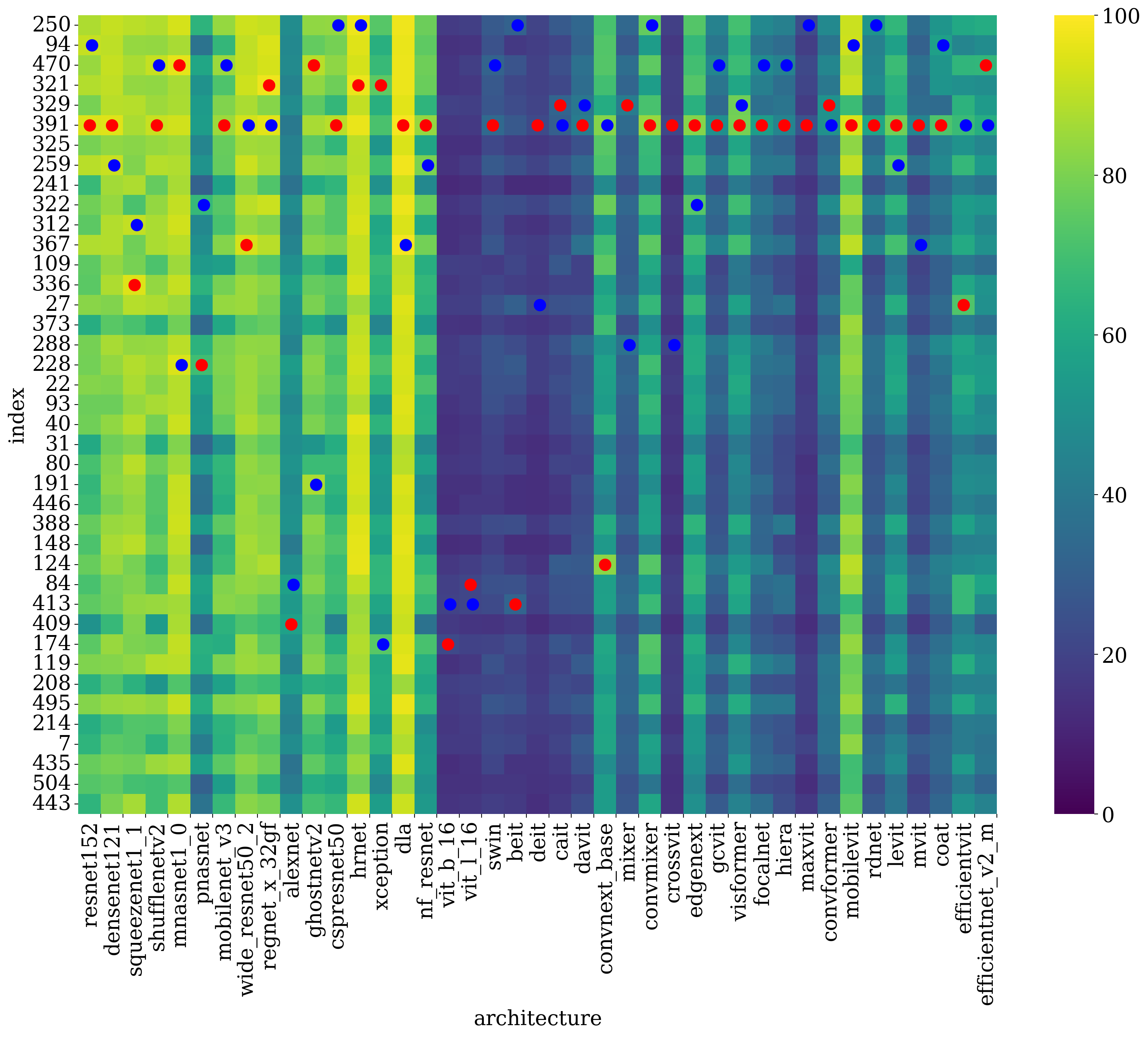}
\caption{{\bf Generator's Performance in NAT.} We plot the heatmap of fooling rates for top-$k=40$ generators across 41 ImageNet target models. The generators are ordered based on the ranking determined by the DenseNet121~\cite{densenet} held-out model after lightweight training, while the $x$-axis shows the labels of the target models, sorted by the three categories. We highlight the best and second best performing generator for each target model with the red and blue circles, respectively.}
\label{fig:generator_performance}
\end{figure}

\noindent{\bf Insights on Generators.} In Figure~\ref{fig:generator_performance}, we visualize the heatmap of fooling rates for the top-40 generators trained using our NAT approach. On the $y$-axis, generators are ordered based on the ranking determined by the DenseNet121~\cite{densenet} held-out model, and we also show the corresponding attacked Neuron position $j$ for each generator. For example,  in our experiments, the first query uses the adversarial image generated by attacking Neuron position $j=250$, followed by a query with the generator attacking Neuron $j=94$, and so on. 

We observe that, in general, all generators exhibit higher transferability on convolutional models. Interestingly, the generator $G_{321}$ attacking Neuron position $j=391$ performs best on 23 out of 41 target models (indicated with red markers). For the remaining 18 models, the generator $G_{321}$ attacking Neuron position $j=321$ performs best in 3 cases, while the top-performing generators for the other models are distributed across different neuron positions. This plot also highlights that no single generator consistently outperforms the others across all target models, underscoring the importance of employing multiple diverse adversarial patterns for a successful attack, depending on the target model architecture. In contrast, existing methods like LTP and BIA primarily attack neurons tied to a single concept (closely resembling Neuron $j=241$), limiting their transferability.

\begin{table}[!htb]
\centering

\resizebox{0.8\columnwidth}{!}{
\begin{subtable}{\columnwidth}
    \centering
    \begin{tabular}{cccc}
    \toprule
    \multirow{2}{*}{\textbf{Held-out}} & \multicolumn{3}{c}{\textbf{Cross-Model Results}}  \\
    \cmidrule(lr){2-4}
     & \textbf{Conv.} $\uparrow$ & \textbf{Transformer} $\uparrow$ & \textbf{Mixed}  $\uparrow$  \\ \midrule
    DenseNet121~\cite{densenet} & 96.2 & 53.3 & 79.0 \\ 
    ResNet152~\cite{resnet}   & 96.1 & 54.4 & 79.4 \\  
    VGG16~\cite{vgg}       & 95.6 & 51.4 & 78.3 \\ 
    \bottomrule
    \end{tabular}
    \caption{Cross-model results}
    \label{tab:1a}
\end{subtable}}

\vspace{0.5cm} 

\resizebox{0.7\columnwidth}{!}{
\centering
\begin{subtable}{\columnwidth}
    \centering
    \begin{tabular}{ccccc}
    \toprule
    \multirow{2}{*}{\textbf{Held-out}} & \multicolumn{4}{c}{\textbf{Cross-Domain Results}}  \\
    \cmidrule(lr){2-5}
     & \textbf{CUB} $\downarrow$ & \textbf{S-Cars} $\downarrow$ & \textbf{FGVC} $\downarrow$ & \textbf{Avg.} $\downarrow$  \\ \midrule
    DenseNet121~\cite{densenet} & 14.8 & 7.9 & 18.5 & 13.6 \\ 
    ResNet152~\cite{resnet}   & 14.8 & 7.6 & 6.26 & 9.5 \\  
    VGG16~\cite{vgg}       & 16.9 & 7.9 & 19.1 & 14.6 \\ 
    \bottomrule
    \end{tabular}
    \caption{Cross-domain results}
    \label{tab:1b}
\end{subtable}}

\caption{{\bf Ablation with Different Held-out Models for Top-k Neuron Selection.} The results for cross-model (top) and cross-domain (bottom) settings are reported with $k=40$ queries.}
\label{tab:ablate_held_out}
\end{table}

\subsubsection{Limitations} Our study focuses on disrupting neurons at a single layer, $l=18$, within the VGG16 model, as this layer was identified as optimal in LTP~\cite{ltp}. However, due to the significant computational costs associated with training individual generators, we could not extend experiments to other layers. Additionally, even for the attacked layer $l=18$, we only targeted a maximum of $40$ neurons out of $512$ possible positions. Future work exploring the identification of both the optimal layer and the optimal set of neurons may reveal even stronger attack capabilities. Another limitation is that, during the selection of the top-$k$ neurons, we observed that some of the generators may target similar concepts, reducing the overall diversity of the attack patterns. This overlap could be avoided by incorporating more refined neuron selection techniques.  Moreover, while our current analysis is limited to a single source model, expanding this framework to include multiple source models could further improve transferability. However, this extension requires significant computational resources, especially when dealing with source models that have a large number of dimensions in their embeddings. Future work could explore efficient attack strategies, potentially combining features from different models to increase attack diversity without dramatically increasing computational costs.

\section{Conclusion}
In this work, we identified that prior generative attacks relying on layer-level embedding separation predominantly focused on disrupting a few neurons associated with a similar concept. Building on this observation, we propose NAT, training mutiple expert generators explicitly designed to target individual neuron, rather than attacking all neurons simultaneously.

Our results show that this targeted approach yields more transferable generators compared to baseline methods due to their specialized focus. Additionally, by querying multiple adversarial images from different generators, we demonstrate that the fooling rate improves significantly within fewer than 10 queries. This highlights that the generators are complementary, effectively disrupting key fundamental units across over 40 ImageNet models and 9 fine-grained dataset models. For future work, we plan to explore ways to further enhance transferability with fewer queries, such as by pruning the trained generators or employing a gating mechanism.

{
 \small
\bibliographystyle{ieee_fullname}
\bibliography{egbib}
}

\clearpage

\section{\bf Implementation}

We provide below additional details for the reproducibility of our experiments.\\

\noindent {\bf Libraries.} We conducted our experiments on NVIDIA GeForce RTX 3090  using PyTorch 2.1.1~\cite{paszke2019pytorch}, CUDA 11.8, Timm 1.0.9~\cite{rw2019timm}, and Torchvision 0.16.1~\cite{torchvision2016}.  \\

\noindent {\bf Generator Modifications.} Due to the non-deterministic behavior\footnote{\href{https://github.com/pytorch/pytorch/issues/98925}{https://github.com/pytorch/pytorch/issues/98925}} of the \texttt{ReflectionPad2d} operation, which was used inside the generator architecture in prior works~\cite{ltp, bia, cda}, we removed it in the \texttt{block1} layer and set padding to 1 in the subsequent \texttt{conv2d} operation within the same layer. We also removed \texttt{ReflectionPad2d} in the \texttt{blockf} layer and set the padding to 5 in the \texttt{conv2d} operation. Furthermore, the \texttt{residualblock} implementation contains two instances of \texttt{ReflectionPad2d}, both of which were removed, with padding set to 1 in the subsequent \texttt{conv2d} operations. These changes allowed us to eliminate \texttt{ReflectionPad2d} from the generator while compensating for changes in feature size with additional padding in the subsequent convolution operations. \\

\begin{figure*}[ht!]
    \centering
    \captionsetup{justification=centering, margin=1cm}
    \begin{subfigure}[b]{0.45\textwidth}
        \centering
        \includegraphics[width=\textwidth]{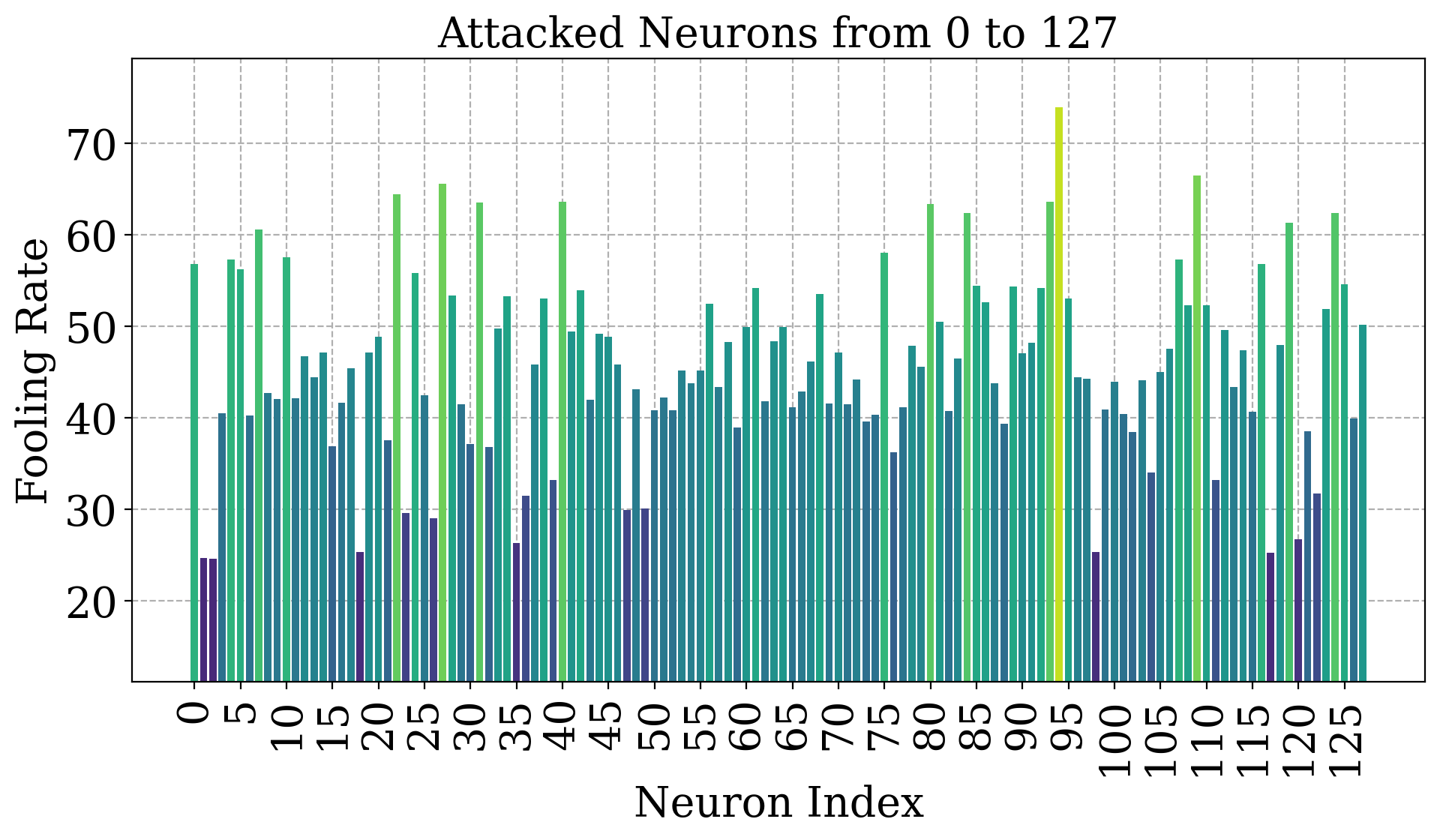}
    \end{subfigure}
        \begin{subfigure}[b]{0.45\textwidth}
        \centering
        \includegraphics[width=\textwidth]{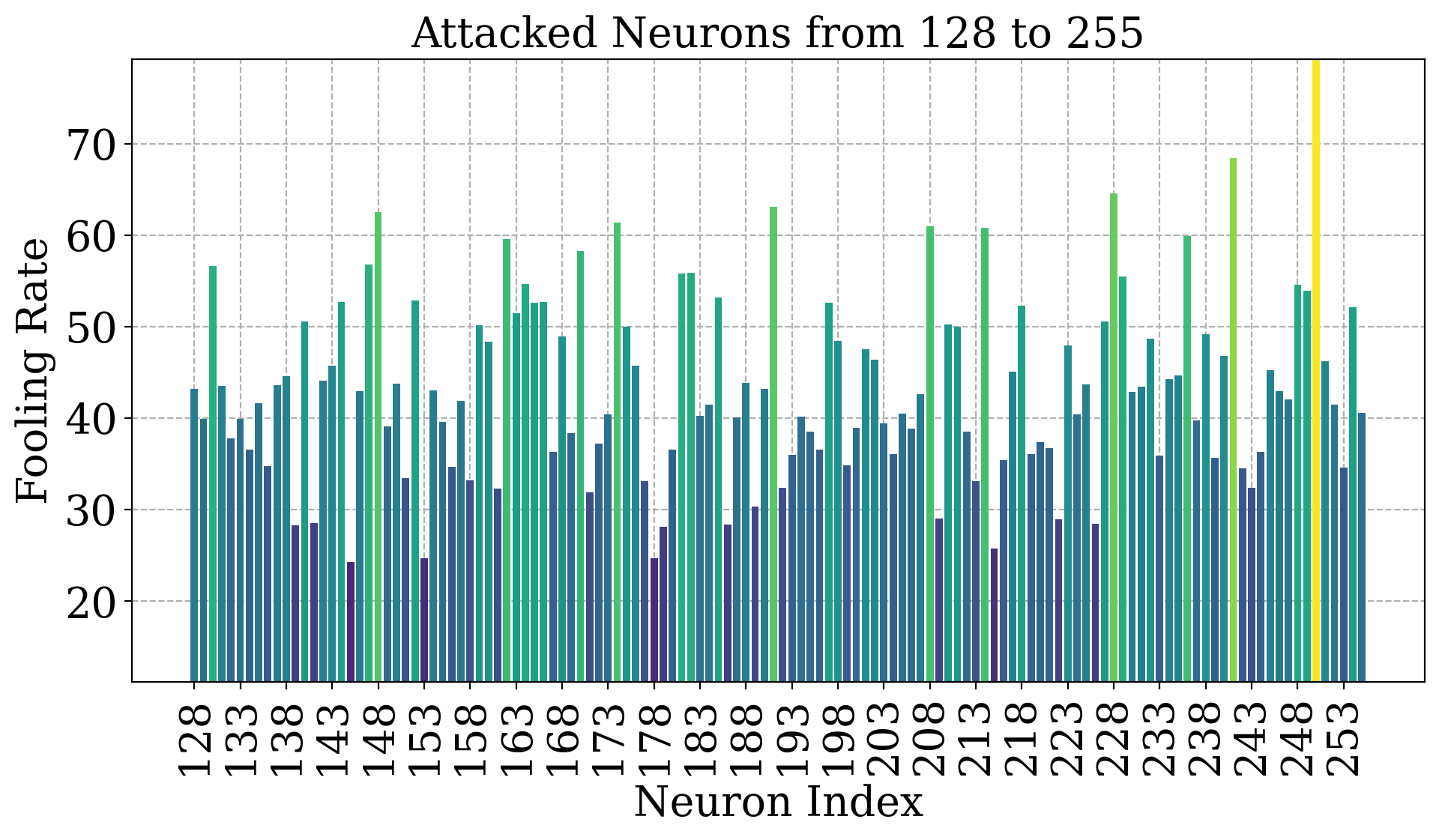}
    \end{subfigure}\\
        \begin{subfigure}[b]{0.45\textwidth}
        \centering
        \includegraphics[width=\textwidth]{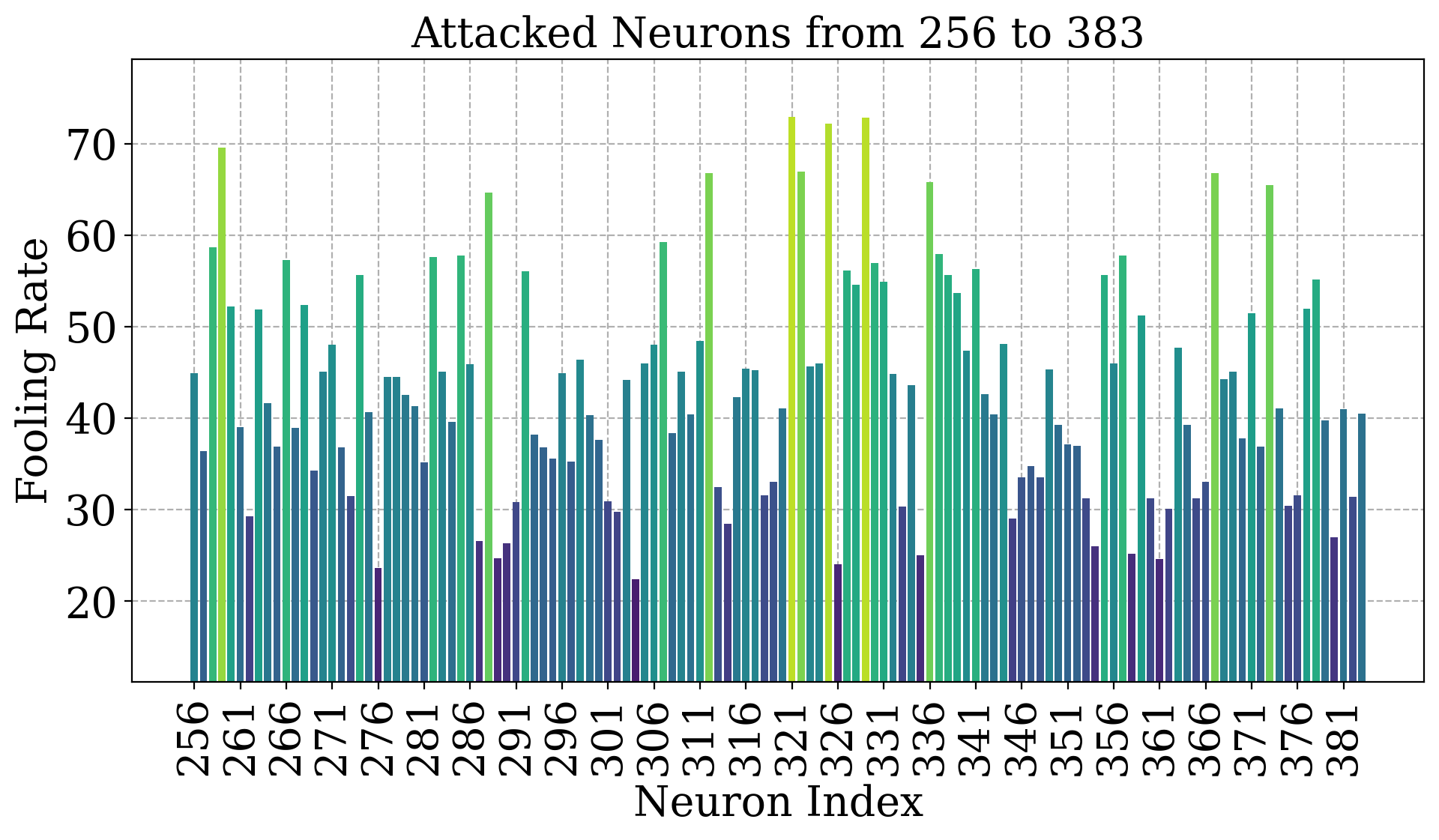}
    \end{subfigure}
        \begin{subfigure}[b]{0.45\textwidth}
        \centering
        \includegraphics[width=\textwidth]{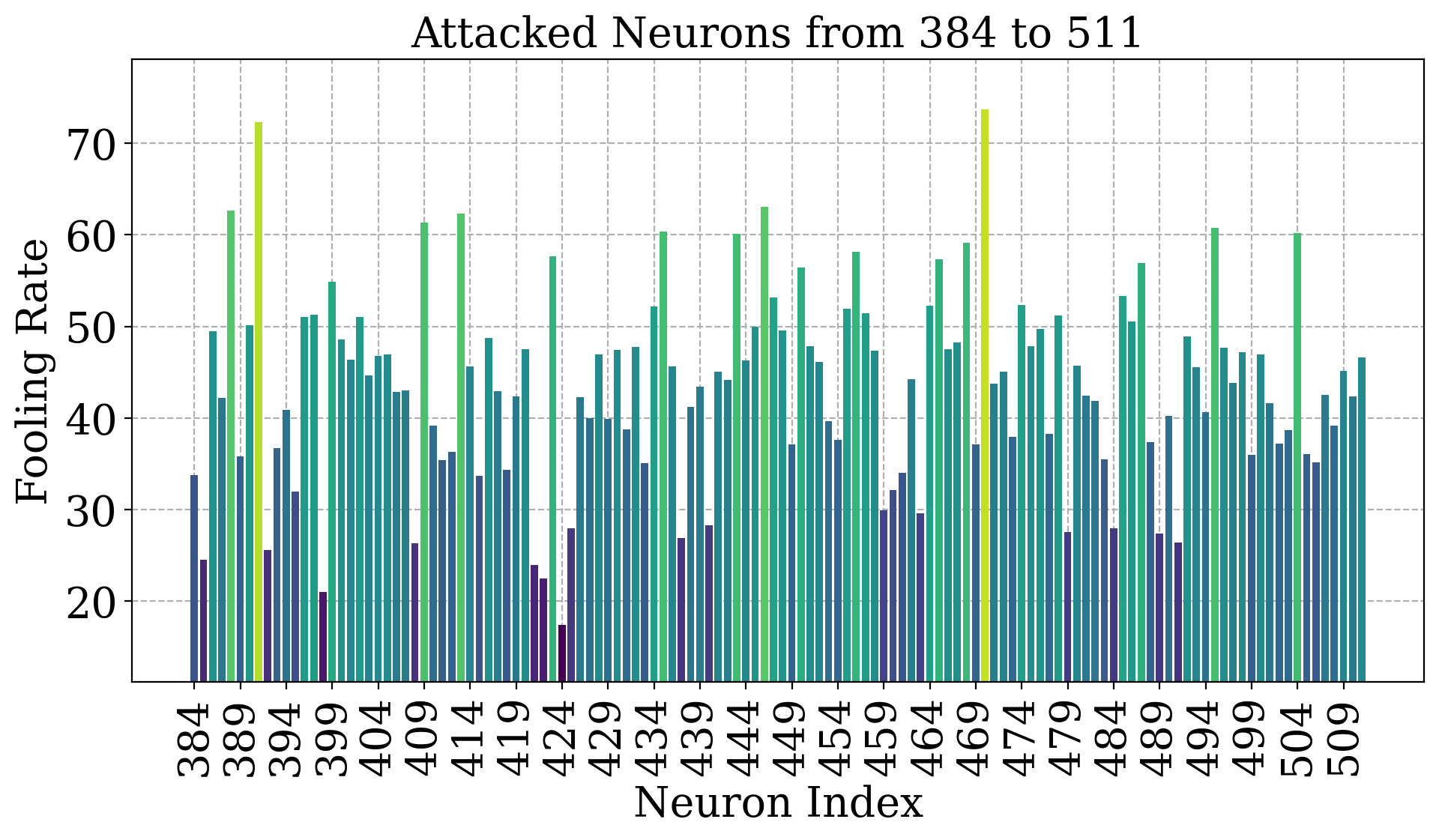}
    \end{subfigure}
\caption{{\bf Evaluation of 512 Neuron specific generators on DenseNet121~\cite{densenet} obtained after end of lightweight training on VGG16~\cite{vgg}}. We observe that some filters are more transferable than others and we choose top-$k$ generators that has highest transferability.}\label{fig:512ondensenet121}
\end{figure*}

\begin{figure*}[ht!]
    \centering
    \captionsetup{justification=centering, margin=1cm}
    \begin{subfigure}[b]{0.45\textwidth}
        \centering
        \includegraphics[width=\textwidth]{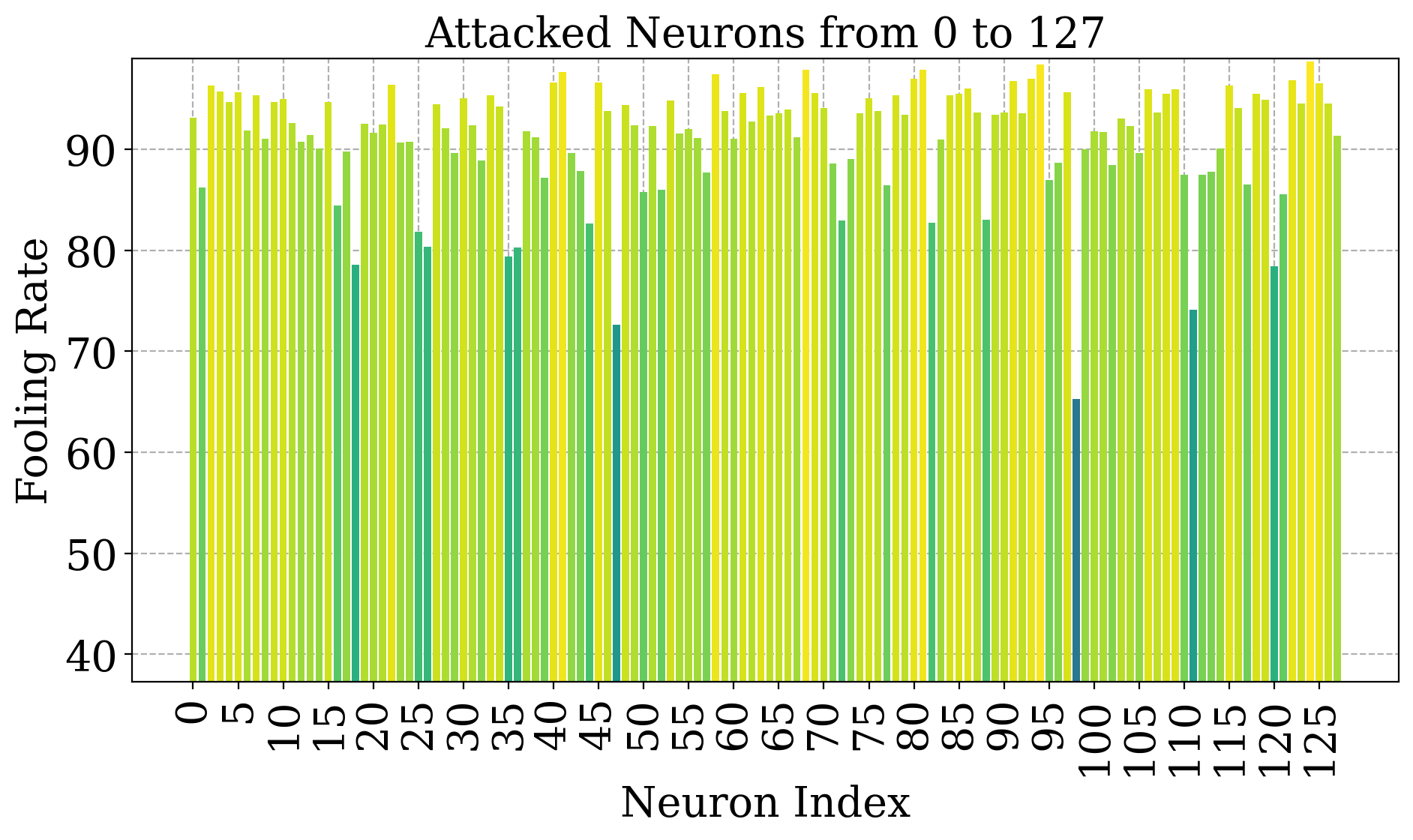}
    \end{subfigure}
        \begin{subfigure}[b]{0.45\textwidth}
        \centering
        \includegraphics[width=\textwidth]{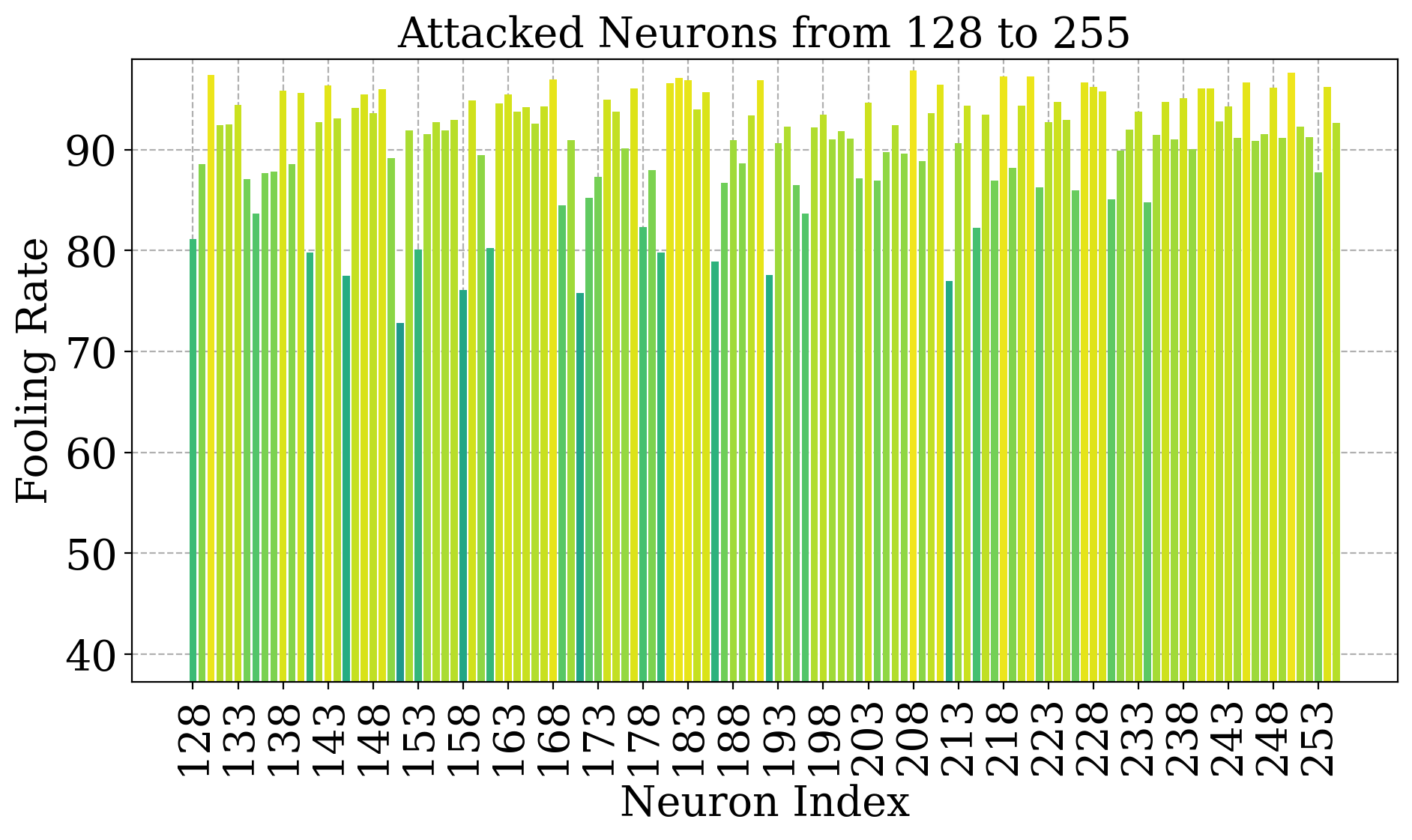}
    \end{subfigure}\\
        \begin{subfigure}[b]{0.45\textwidth}
        \centering
        \includegraphics[width=\textwidth]{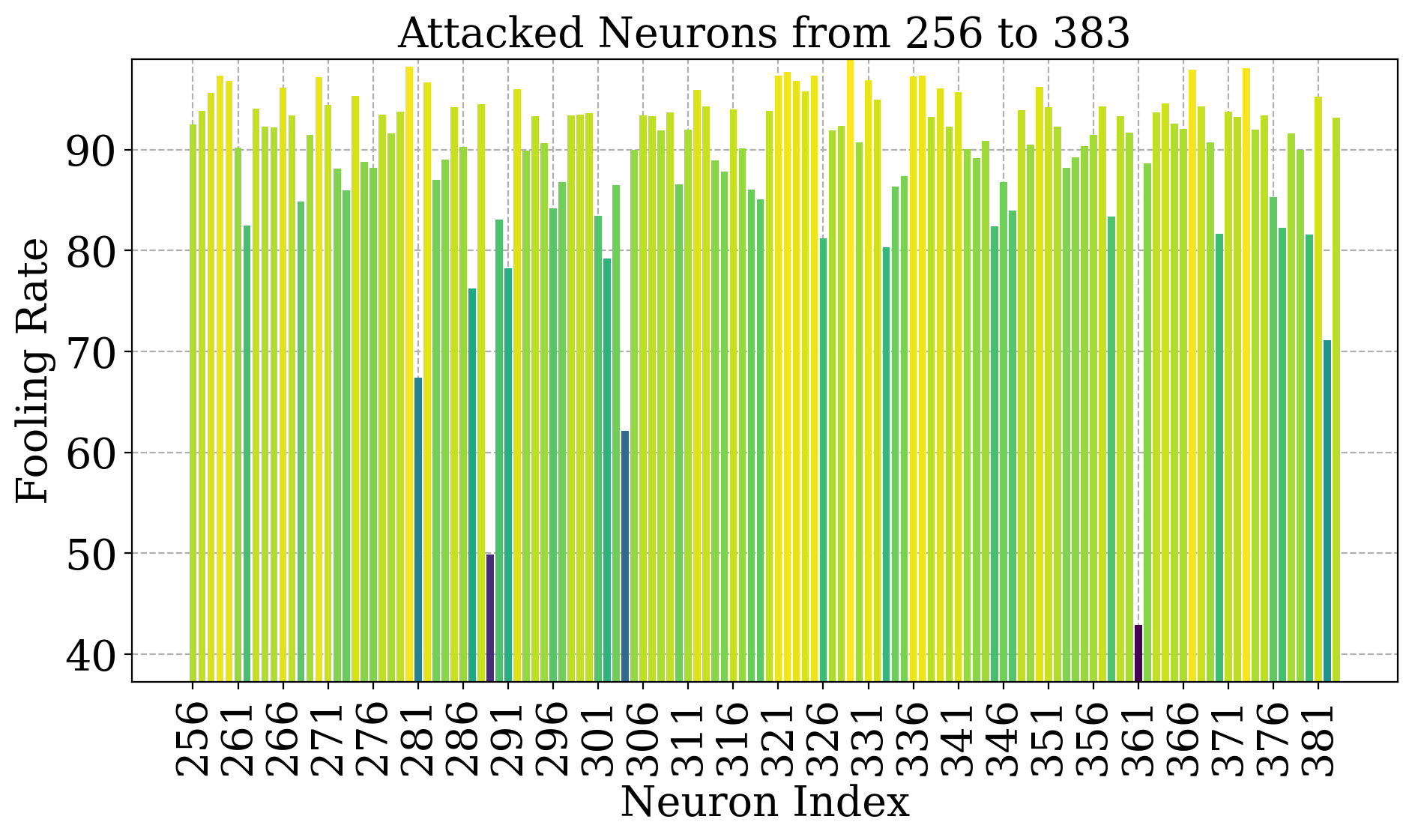}
    \end{subfigure}
        \begin{subfigure}[b]{0.45\textwidth}
        \centering
        \includegraphics[width=\textwidth]{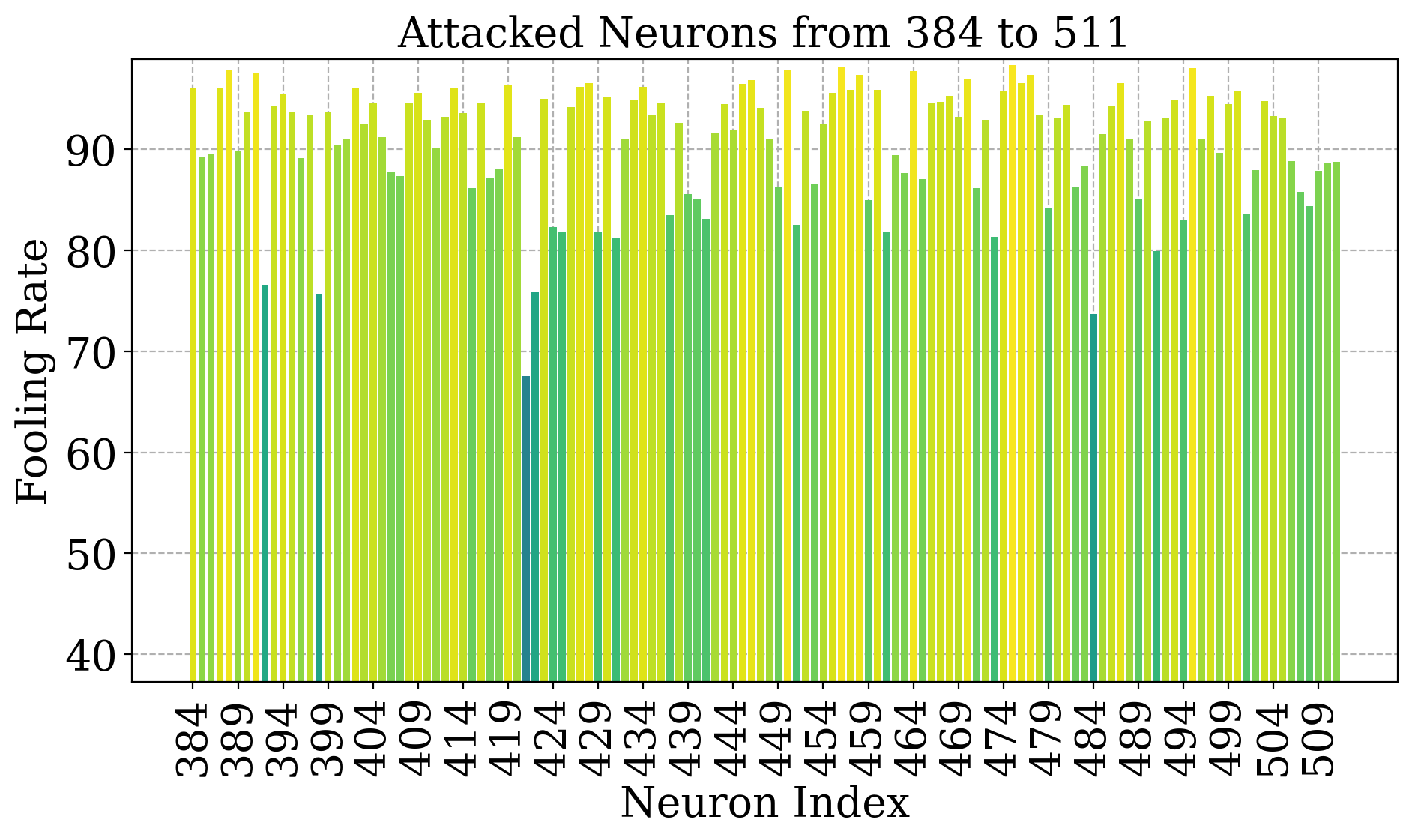}
    \end{subfigure}
\caption{{\bf Evaluation of 512 Neuron specific generators on VGG16~\cite{vgg} obtained after end of lightweight training on the same VGG16~\cite{vgg}.} We observe that all generators have high fooling rates suggesting that overfitting to the source VGG16~\cite{vgg} model.} \label{fig:512onvgg16}
\end{figure*}

\noindent {\bf Top-$k$ Neurons After Initial Lightweight Training.} In Section 3.1, we outlined that all 512 generators, corresponding to the 512 neurons in layer 18 of VGG16, undergo an initial lightweight training phase using 3.125\% of the training set. After this phase, we selected the top-$k=40$ generators based on their performance, either on the source VGG16~\cite{vgg} or by using a held-out model such as DenseNet121~\cite{densenet} or ResNet~\cite{resnet} in our experiments. We provide the top-ranked neurons from this step across different held-out models in Table~\ref{tab:top40neurons}. Additionally, for full transparency, we present the performance of all 512 neurons after the lightweight training phase on two heldout models in Figures~\ref{fig:512ondensenet121}, and \ref{fig:512onvgg16} for DenseNet121~\cite{densenet} and VGG16~\cite{vgg} heldout models. \\

\noindent{\bf Target Models.} We used 41 target models, publicly available in Torchvision~\cite{torchvision2016} and Timm libraries~\cite{rw2019timm}. We provide in Table~\ref{tab:targetmodels} the exact version details used for each model and also report the clean accuracy on the evaluation set of 5K images~\cite{ltp}, sampled from the ImageNet validation set~\cite{imagenet}.  We used 9 models for cross-domain provided by the authors of BIA~\cite{bia}

\section{Additional Quantitative Results}

\noindent {\bf Performance of each neuron-specific generator.} In Tables~\ref{tab:top20genonallmodels} and ~\ref{tab:top20to40genonallmodels}, we provide the performance of all fully trained top-40 generators selected using DenseNet121~\cite{densenet} as the held-out model. We observed that our neuron-specific generators outperform the baselines in 37 out of 40 cases in the single-query setting. Moreover, the best-performing generator $G_{391}$, attacking neuron 391, outperforms the highest-ranked generator $G_{250}$ obtained at the end of lightweight training by more than 4\%. We believe that with better neuron ranking schemes, the fooling rate in the single-query setting can be further enhanced. \\

\begin{figure*}[ht!]
    \centering
    \captionsetup{justification=centering, margin=1cm}
    \begin{subfigure}[b]{0.45\textwidth}
        \centering
        \includegraphics[width=\textwidth]{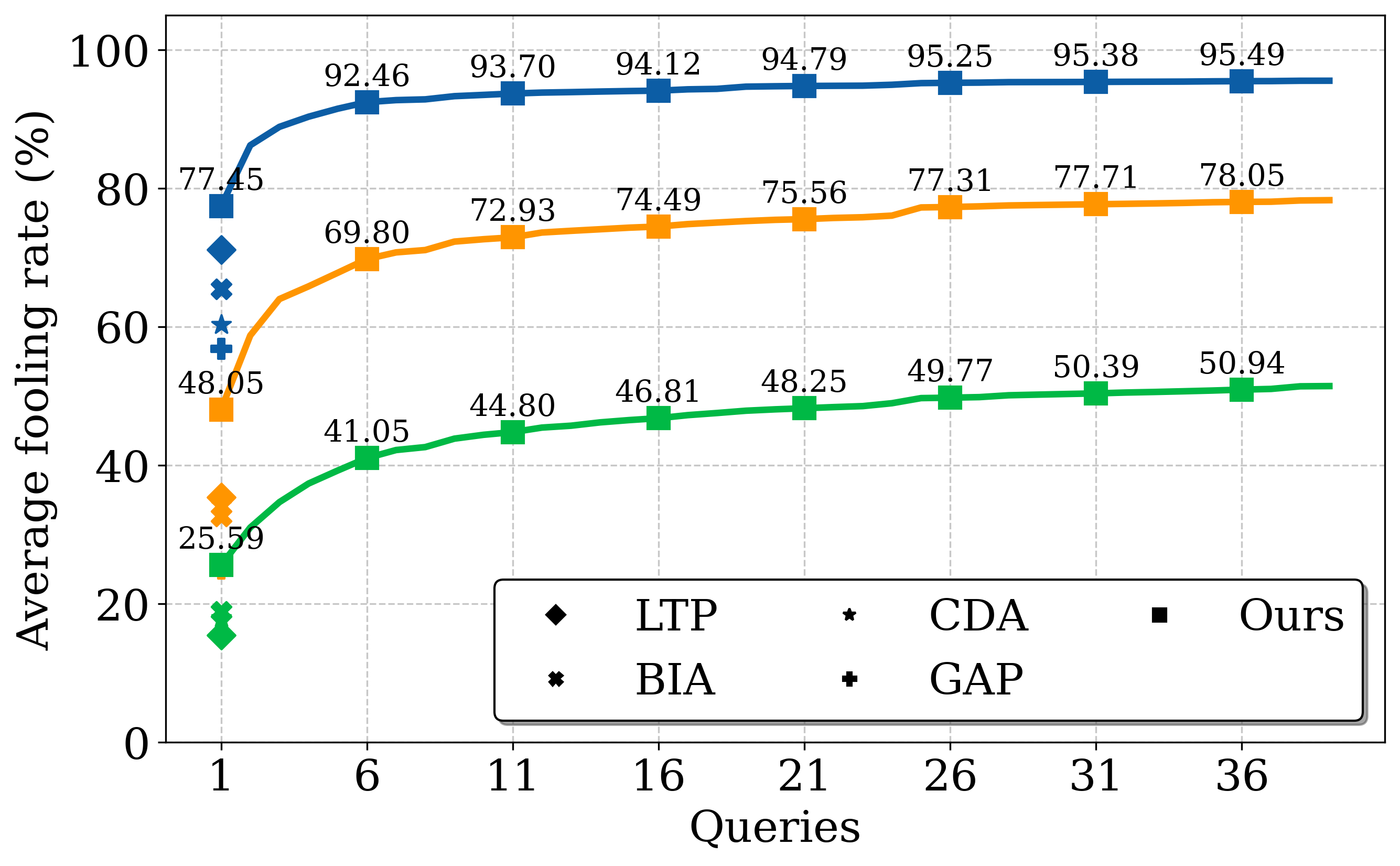}
        \caption*{(a) VGG16~\cite{vgg}}
    \end{subfigure}
        \begin{subfigure}[b]{0.45\textwidth}
        \centering
        \includegraphics[width=\textwidth]{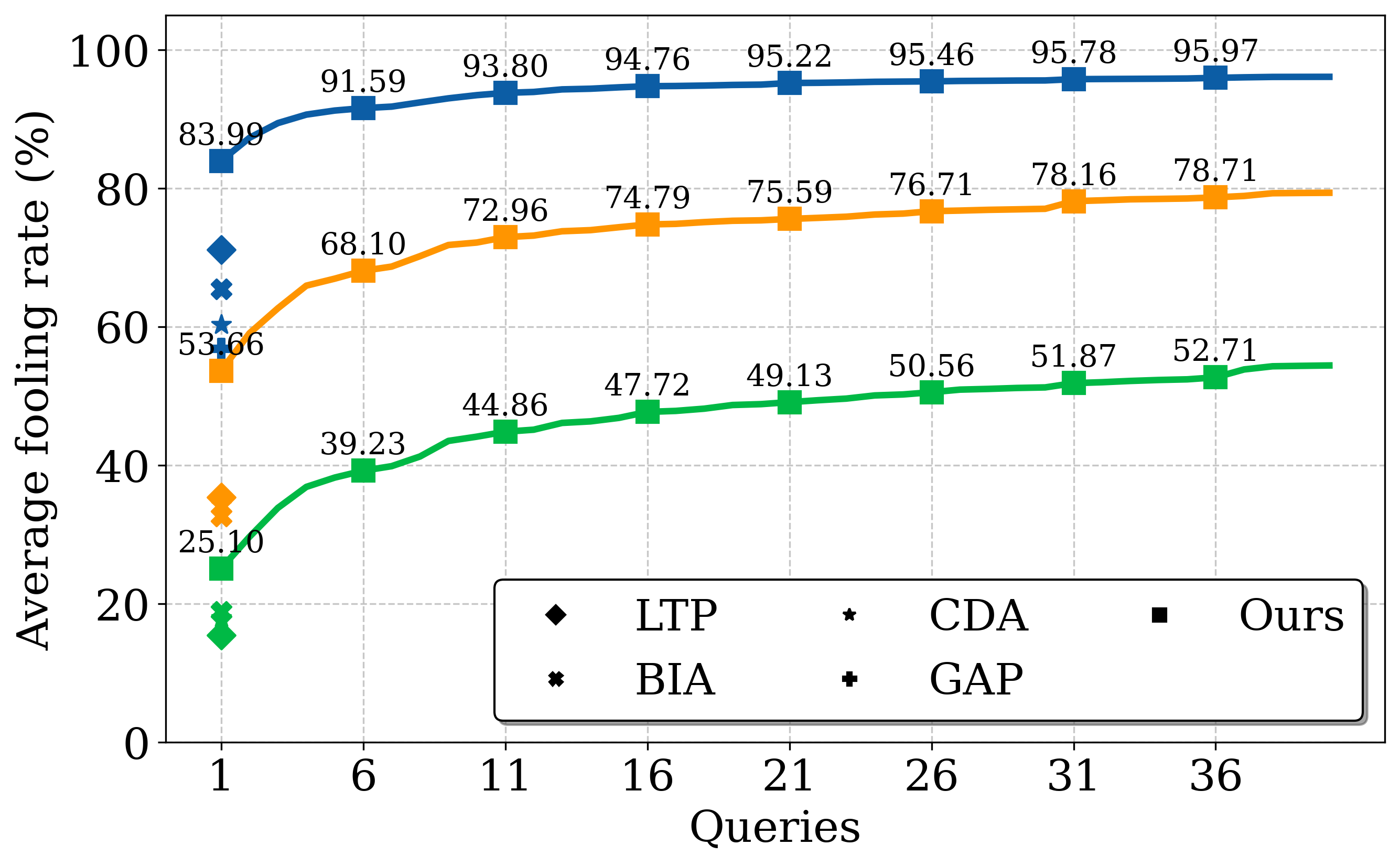}
        \caption*{(b) ResNet152~\cite{resnet}}
    \end{subfigure}\\
\caption{{\bf Impact of heldout model for top-k neuron selection after lightweight training} We use the same source model VGG16~\cite{vgg} as the heldout model for top-$k$ neuron selection after lightweight training in the left and similarily using ResNet152~\cite{resnet} on the right. We observe that for both the cases, the fooling rate with k-queries differ marginally and perform along the same lines as reported in the main paper with DenseNet121~\cite{densenet} model.}\label{fig:query_vs_performance_supp}
\end{figure*}

\noindent {\bf Choice of Heldout model for top-$k$ neuron selection.} In Figure~\ref{fig:query_vs_performance_supp}, we show the performance of top-40 generators trained with different models for neuron selection versus the number of queries. We observe that our top-k neuron position is robust to the choice of heldout model selection. And in all cases, top-1 generator consisted outperformed baselines in the single-query setting.

\section{Additional Qualitative Results}
\begin{figure*}[ht!]
    \centering
    \captionsetup{justification=centering, margin=1cm}
    \begin{subfigure}[b]{0.22\textwidth}
        \centering
        \includegraphics[width=\textwidth]{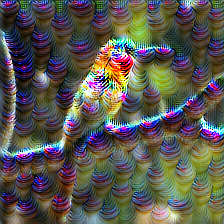}
        \caption*{Adversarial Image, Neuron: 250}
    \end{subfigure}
    \begin{subfigure}[b]{0.22\textwidth}
        \centering
    \includegraphics[width=\textwidth]{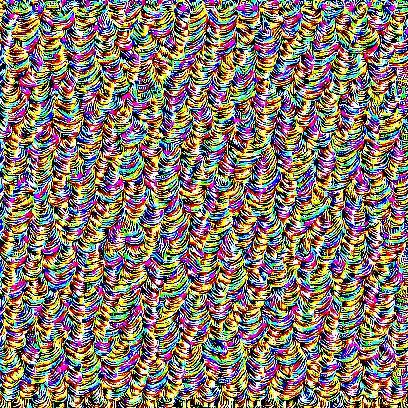}
        \caption*{Synthesized Image, Neuron: 250}
    \end{subfigure}
    \begin{subfigure}[b]{0.22\textwidth}
        \centering
        \includegraphics[width=\textwidth]{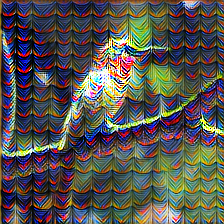}
        \caption*{Synthesized Image, Neuron: 94}
    \end{subfigure}
    \begin{subfigure}[b]{0.22\textwidth}
        \centering
    \includegraphics[width=\textwidth]{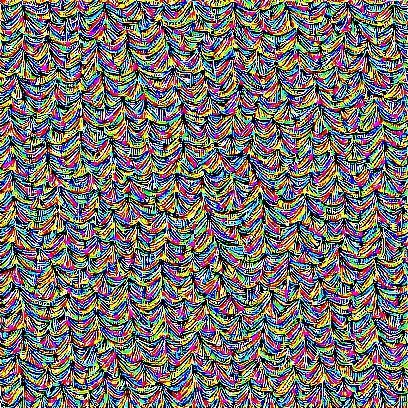}
        \caption*{Synthesized Image, Neuron: 94}
    \end{subfigure}
    \\
    \begin{subfigure}[b]{0.22\textwidth}
        \centering
        \includegraphics[width=\textwidth]{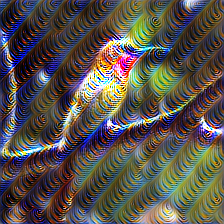}
        \caption*{Adversarial Image, Neuron: 470}
    \end{subfigure}
    \begin{subfigure}[b]{0.22\textwidth}
        \centering
    \includegraphics[width=\textwidth]{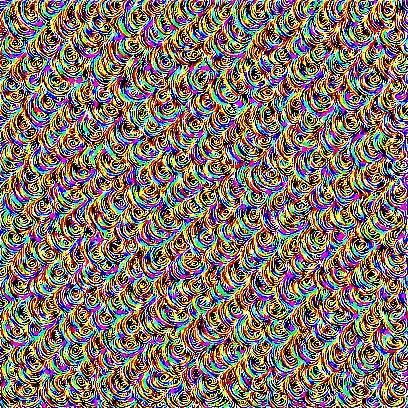}
        \caption*{Synthesized Image, Neuron: 470}
    \end{subfigure}
    \begin{subfigure}[b]{0.22\textwidth}
        \centering
        \includegraphics[width=\textwidth]{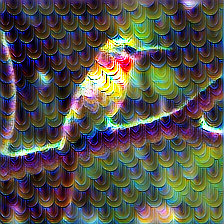}
        \caption*{Synthesized Image, Neuron: 321}
    \end{subfigure}
    \begin{subfigure}[b]{0.22\textwidth}
        \centering
    \includegraphics[width=\textwidth]{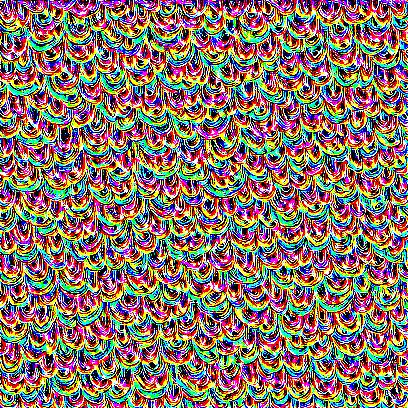}
        \caption*{Synthesized Image, Neuron: 321}
    \end{subfigure}
    \\
     \begin{subfigure}[b]{0.22\textwidth}
        \centering
        \includegraphics[width=\textwidth]{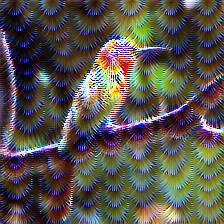}
        \caption*{Adversarial Image, Neuron: 329}
    \end{subfigure}
    \begin{subfigure}[b]{0.22\textwidth}
        \centering
    \includegraphics[width=\textwidth]{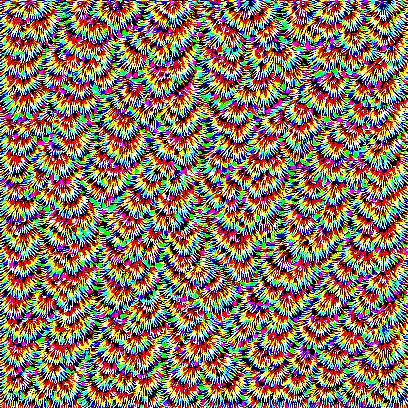}
        \caption*{Synthesized Image, Neuron: 329}
    \end{subfigure}
    \begin{subfigure}[b]{0.22\textwidth}
        \centering
        \includegraphics[width=\textwidth]{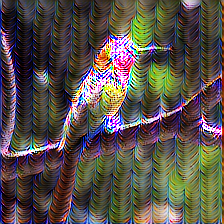}
        \caption*{Synthesized Image, Neuron: 391}
    \end{subfigure}
    \begin{subfigure}[b]{0.22\textwidth}
        \centering
    \includegraphics[width=\textwidth]{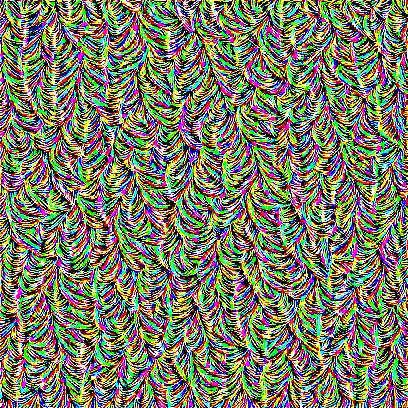}
        \caption*{Synthesized Image, Neuron: 391}
    \end{subfigure}
    \\
    \begin{subfigure}[b]{0.22\textwidth}
        \centering
        \includegraphics[width=\textwidth]{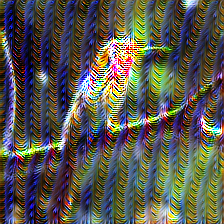}
        \caption*{Adversarial Image, Neuron: 325}
    \end{subfigure}
    \begin{subfigure}[b]{0.22\textwidth}
        \centering
    \includegraphics[width=\textwidth]{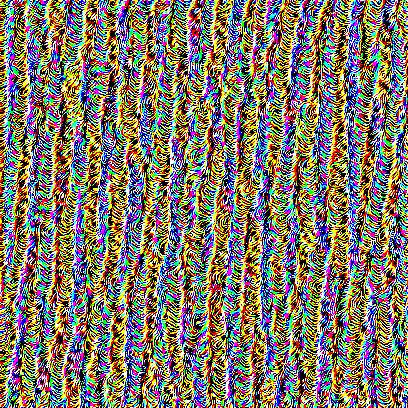}
        \caption*{Synthesized Image, Neuron: 325}
    \end{subfigure}
    \begin{subfigure}[b]{0.22\textwidth}
        \centering
        \includegraphics[width=\textwidth]{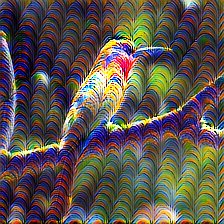}
        \caption*{Synthesized Image, Neuron: 259}
    \end{subfigure}
    \begin{subfigure}[b]{0.22\textwidth}
        \centering
    \includegraphics[width=\textwidth]{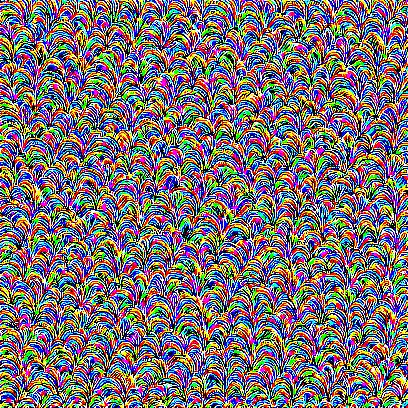}
        \caption*{Synthesized Image, Neuron: 259}
    \end{subfigure}
    \\
    \begin{subfigure}[b]{0.22\textwidth}
        \centering
        \includegraphics[width=\textwidth]{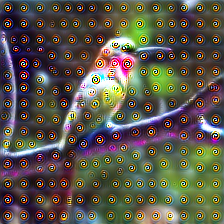}
        \caption*{Adversarial Image, Neuron: 241}
    \end{subfigure}
    \begin{subfigure}[b]{0.22\textwidth}
        \centering
    \includegraphics[width=\textwidth]{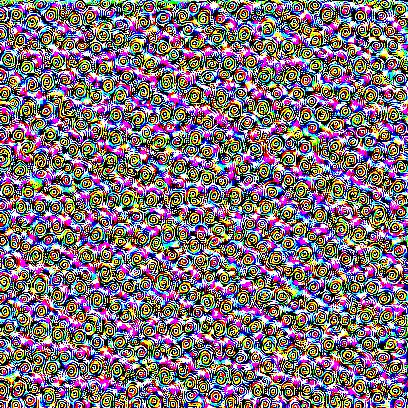}
        \caption*{Synthesized Image, Neuron: 241}
    \end{subfigure}
    \begin{subfigure}[b]{0.22\textwidth}
        \centering
        \includegraphics[width=\textwidth]{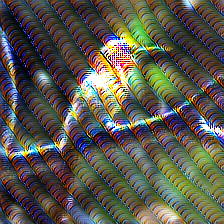}
        \caption*{Synthesized Image, Neuron: 322}
    \end{subfigure}
    \begin{subfigure}[b]{0.22\textwidth}
        \centering
    \includegraphics[width=\textwidth]{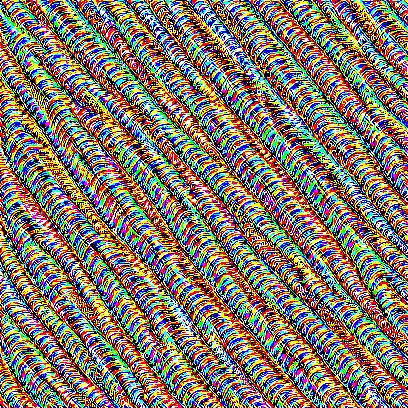}
        \caption*{Synthesized Image, Neuron: 322}
    \end{subfigure}
    \\
\caption{Generated unbounded adversarial images along with synthesized neuron visualizations for the top 10 attacked neurons. The positions of the neurons are listed below each image. Note that, the top-k neurons were selected based on the transferability to the  DenseNet121~\cite{densenet} as the held-out model after initial lightweight training.}
\label{fig:vis1}
\end{figure*}

\begin{figure*}[ht!]
    \centering
    \captionsetup{justification=centering, margin=1cm}
    \begin{subfigure}[b]{0.22\textwidth}
        \centering
        \includegraphics[width=\textwidth]{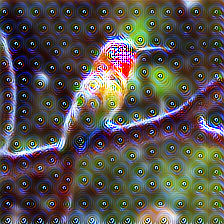}
        \caption*{Adversarial Image, Neuron: 312}
    \end{subfigure}
    \begin{subfigure}[b]{0.22\textwidth}
        \centering
    \includegraphics[width=\textwidth]{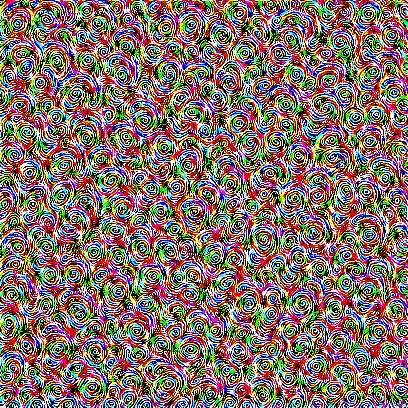}
        \caption*{Synthesized Image, Neuron: 312}
    \end{subfigure}
    \begin{subfigure}[b]{0.22\textwidth}
        \centering
        \includegraphics[width=\textwidth]{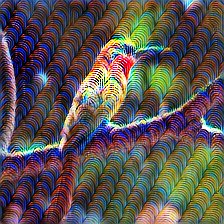}
        \caption*{Synthesized Image, Neuron: 367}
    \end{subfigure}
    \begin{subfigure}[b]{0.22\textwidth}
        \centering
    \includegraphics[width=\textwidth]{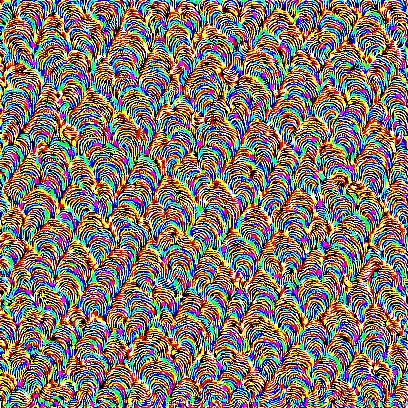}
        \caption*{Synthesized Image, Neuron: 367}
    \end{subfigure}
    \\
    \begin{subfigure}[b]{0.22\textwidth}
        \centering
        \includegraphics[width=\textwidth]{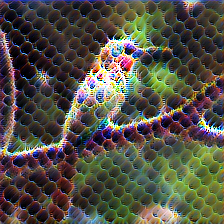}
        \caption*{Adversarial Image, Neuron: 109}
    \end{subfigure}
    \begin{subfigure}[b]{0.22\textwidth}
        \centering
    \includegraphics[width=\textwidth]{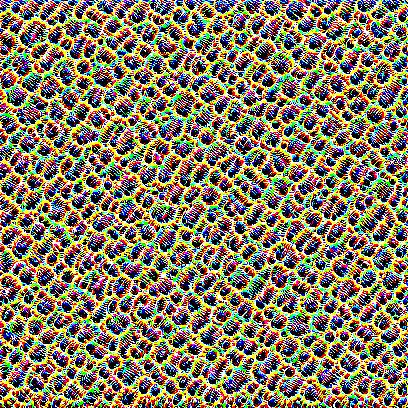}
        \caption*{Synthesized Image, Neuron: 109}
    \end{subfigure}
    \begin{subfigure}[b]{0.22\textwidth}
        \centering
        \includegraphics[width=\textwidth]{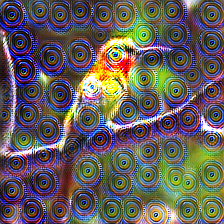}
        \caption*{Synthesized Image, Neuron: 336}
    \end{subfigure}
    \begin{subfigure}[b]{0.22\textwidth}
        \centering
    \includegraphics[width=\textwidth]{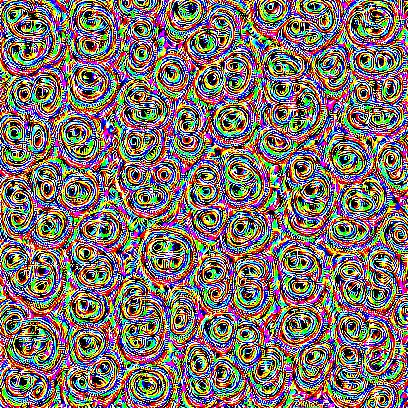}
        \caption*{Synthesized Image, Neuron: 336}
    \end{subfigure}
    \\
     \begin{subfigure}[b]{0.22\textwidth}
        \centering
        \includegraphics[width=\textwidth]{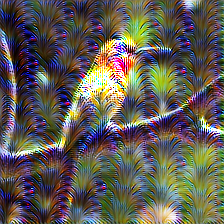}
        \caption*{Adversarial Image, Neuron: 27}
    \end{subfigure}
    \begin{subfigure}[b]{0.22\textwidth}
        \centering
    \includegraphics[width=\textwidth]{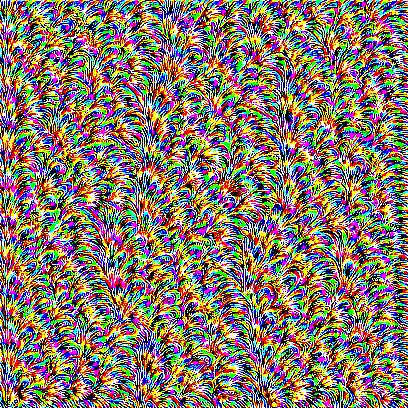}
        \caption*{Synthesized Image, Neuron: 27}
    \end{subfigure}
    \begin{subfigure}[b]{0.22\textwidth}
        \centering
        \includegraphics[width=\textwidth]{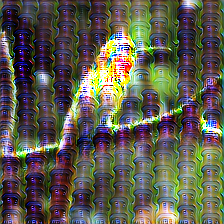}
        \caption*{Synthesized Image, Neuron: 373}
    \end{subfigure}
    \begin{subfigure}[b]{0.22\textwidth}
        \centering
    \includegraphics[width=\textwidth]{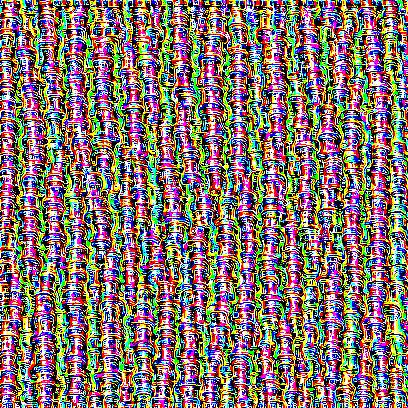}
        \caption*{Synthesized Image, Neuron: 373}
    \end{subfigure}
    \\
    \begin{subfigure}[b]{0.22\textwidth}
        \centering
        \includegraphics[width=\textwidth]{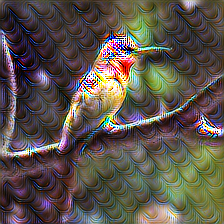}
        \caption*{Adversarial Image, Neuron: 288}
    \end{subfigure}
    \begin{subfigure}[b]{0.22\textwidth}
        \centering
    \includegraphics[width=\textwidth]{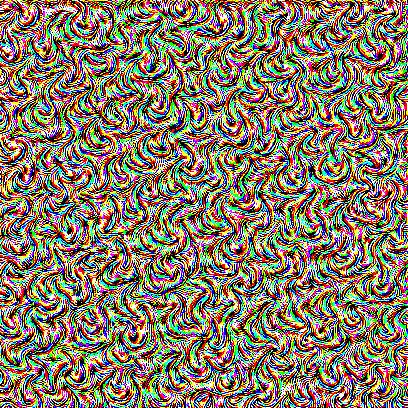}
        \caption*{Synthesized Image, Neuron: 288}
    \end{subfigure}
    \begin{subfigure}[b]{0.22\textwidth}
        \centering
        \includegraphics[width=\textwidth]{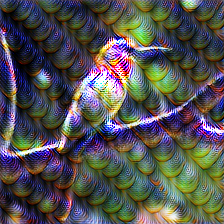}
        \caption*{Synthesized Image, Neuron: 228}
    \end{subfigure}
    \begin{subfigure}[b]{0.22\textwidth}
        \centering
    \includegraphics[width=\textwidth]{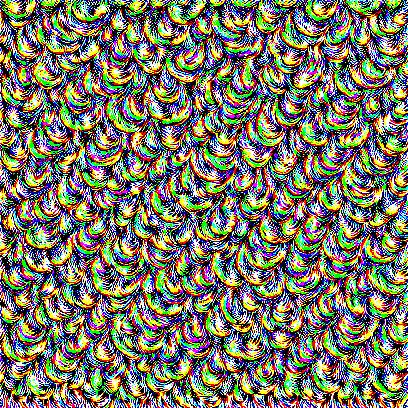}
        \caption*{Synthesized Image, Neuron: 228}
    \end{subfigure}
    \\
    \begin{subfigure}[b]{0.22\textwidth}
        \centering
        \includegraphics[width=\textwidth]{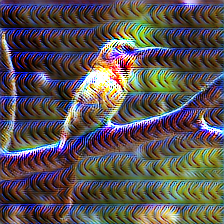}
        \caption*{Adversarial Image, Neuron: 22}
    \end{subfigure}
    \begin{subfigure}[b]{0.22\textwidth}
        \centering
    \includegraphics[width=\textwidth]{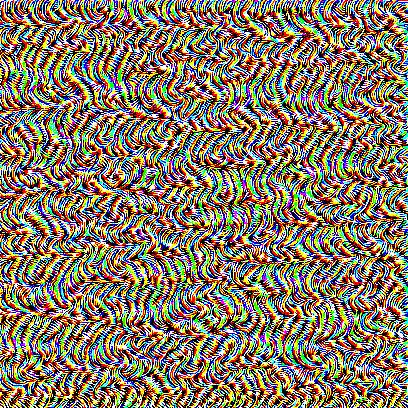}
        \caption*{Synthesized Image, Neuron: 22}
    \end{subfigure}
    \begin{subfigure}[b]{0.22\textwidth}
        \centering
        \includegraphics[width=\textwidth]{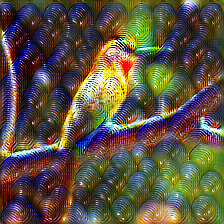}
        \caption*{Synthesized Image, Neuron: 93}
    \end{subfigure}
    \begin{subfigure}[b]{0.22\textwidth}
        \centering
    \includegraphics[width=\textwidth]{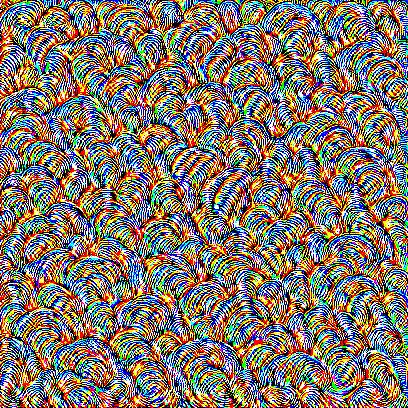}
        \caption*{Synthesized Image, Neuron: 93}
    \end{subfigure}
    \\
\caption{Generated unbounded adversarial images along with synthesized neuron visualizations for the top 10 to 20 attacked neurons. The positions of the neurons are listed below each image. Note that, the top-k neurons were selected based on the transferability to the  DenseNet121~\cite{densenet} as the held-out model after initial lightweight training}
\label{fig:vis2}
\end{figure*}

\begin{figure*}[ht!]
    \centering
    \captionsetup{justification=centering, margin=1cm}
    \begin{subfigure}[b]{0.22\textwidth}
        \centering
        \includegraphics[width=\textwidth]{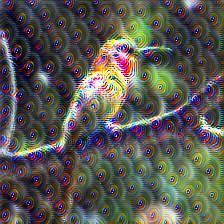}
        \caption*{Adversarial Image, Neuron: 40}
    \end{subfigure}
    \begin{subfigure}[b]{0.22\textwidth}
        \centering
    \includegraphics[width=\textwidth]{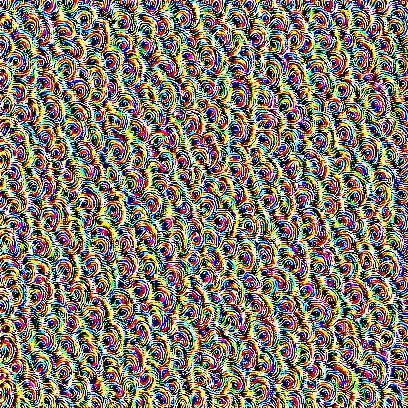}
        \caption*{Synthesized Image, Neuron: 40}
    \end{subfigure}
    \begin{subfigure}[b]{0.22\textwidth}
        \centering
        \includegraphics[width=\textwidth]{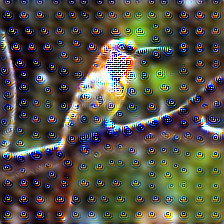}
        \caption*{Synthesized Image, Neuron: 31}
    \end{subfigure}
    \begin{subfigure}[b]{0.22\textwidth}
        \centering
    \includegraphics[width=\textwidth]{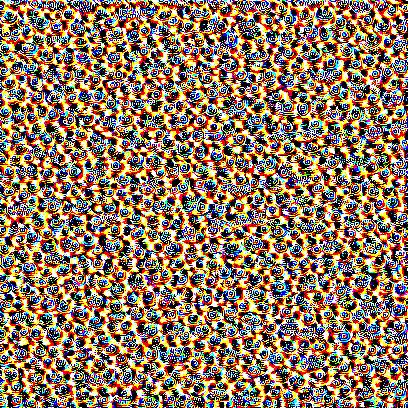}
        \caption*{Synthesized Image, Neuron: 31}
    \end{subfigure}
    \\
    \begin{subfigure}[b]{0.22\textwidth}
        \centering
        \includegraphics[width=\textwidth]{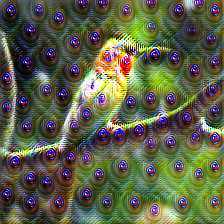}
        \caption*{Adversarial Image, Neuron: 80}
    \end{subfigure}
    \begin{subfigure}[b]{0.22\textwidth}
        \centering
    \includegraphics[width=\textwidth]{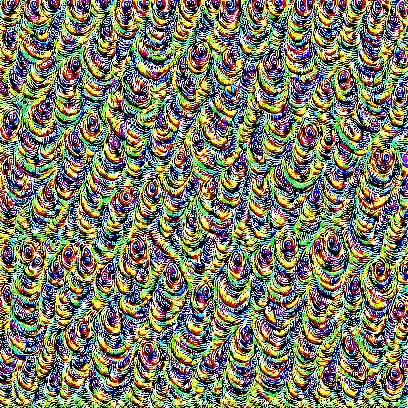}
        \caption*{Synthesized Image, Neuron: 80}
    \end{subfigure}
    \begin{subfigure}[b]{0.22\textwidth}
        \centering
        \includegraphics[width=\textwidth]{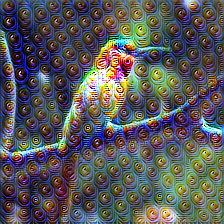}
        \caption*{Synthesized Image, Neuron: 191}
    \end{subfigure}
    \begin{subfigure}[b]{0.22\textwidth}
        \centering
    \includegraphics[width=\textwidth]{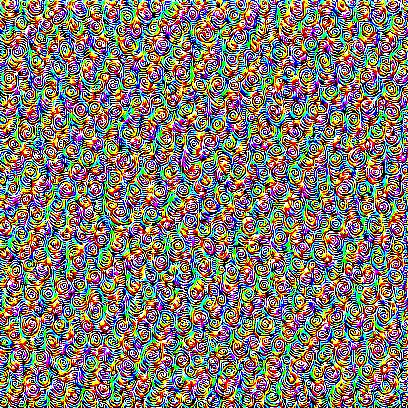}
        \caption*{Synthesized Image, Neuron: 191}
    \end{subfigure}
    \\
     \begin{subfigure}[b]{0.22\textwidth}
        \centering
        \includegraphics[width=\textwidth]{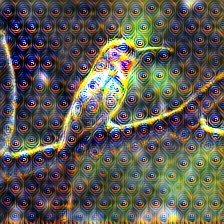}
        \caption*{Adversarial Image, Neuron: 446}
    \end{subfigure}
    \begin{subfigure}[b]{0.22\textwidth}
        \centering
    \includegraphics[width=\textwidth]{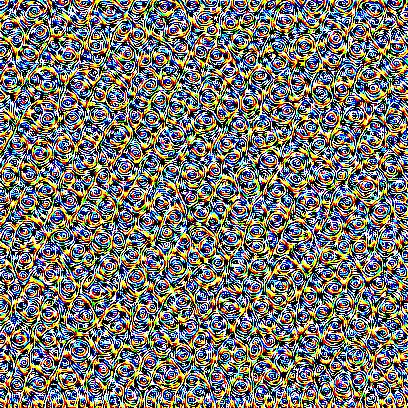}
        \caption*{Synthesized Image, Neuron: 446}
    \end{subfigure}
    \begin{subfigure}[b]{0.22\textwidth}
        \centering
        \includegraphics[width=\textwidth]{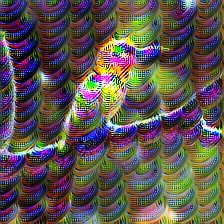}
        \caption*{Synthesized Image, Neuron: 388}
    \end{subfigure}
    \begin{subfigure}[b]{0.22\textwidth}
        \centering
    \includegraphics[width=\textwidth]{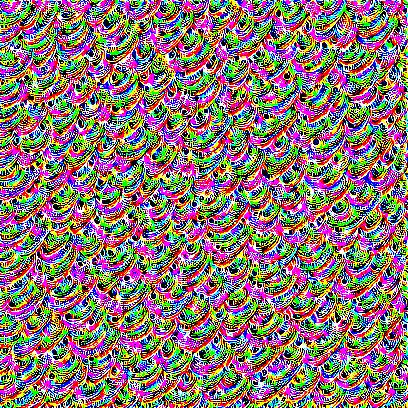}
        \caption*{Synthesized Image, Neuron: 388}
    \end{subfigure}
    \\
    \begin{subfigure}[b]{0.22\textwidth}
        \centering
        \includegraphics[width=\textwidth]{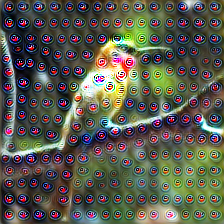}
        \caption*{Adversarial Image, Neuron: 148}
    \end{subfigure}
    \begin{subfigure}[b]{0.22\textwidth}
        \centering
    \includegraphics[width=\textwidth]{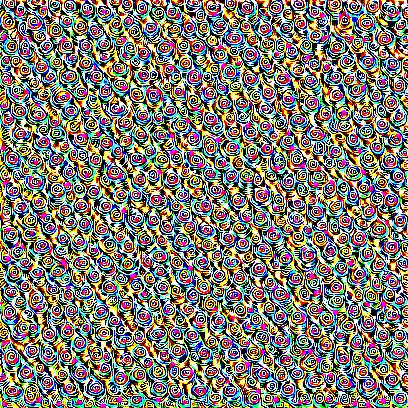}
        \caption*{Synthesized Image, Neuron: 148}
    \end{subfigure}
    \begin{subfigure}[b]{0.22\textwidth}
        \centering
        \includegraphics[width=\textwidth]{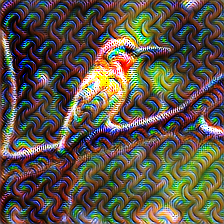}
        \caption*{Synthesized Image, Neuron: 124}
    \end{subfigure}
    \begin{subfigure}[b]{0.22\textwidth}
        \centering
    \includegraphics[width=\textwidth]{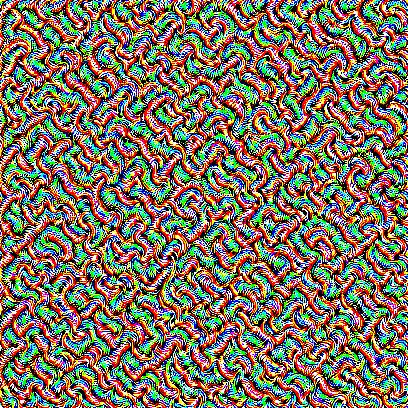}
        \caption*{Synthesized Image, Neuron: 124}
    \end{subfigure}
    \\
    \begin{subfigure}[b]{0.22\textwidth}
        \centering
        \includegraphics[width=\textwidth]{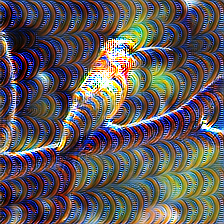}
        \caption*{Adversarial Image, Neuron: 84}
    \end{subfigure}
    \begin{subfigure}[b]{0.22\textwidth}
        \centering
    \includegraphics[width=\textwidth]{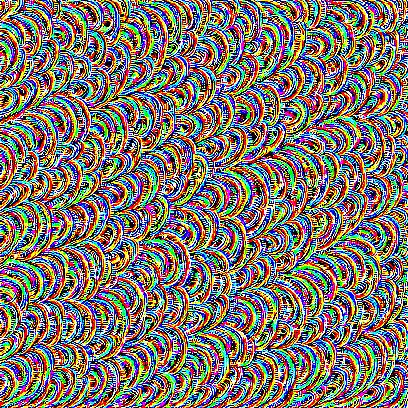}
        \caption*{Synthesized Image, Neuron: 84}
    \end{subfigure}
    \begin{subfigure}[b]{0.22\textwidth}
        \centering
        \includegraphics[width=\textwidth]{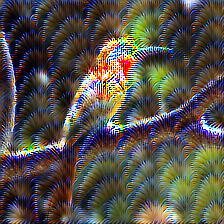}
        \caption*{Synthesized Image, Neuron: 413}
    \end{subfigure}
    \begin{subfigure}[b]{0.22\textwidth}
        \centering
    \includegraphics[width=\textwidth]{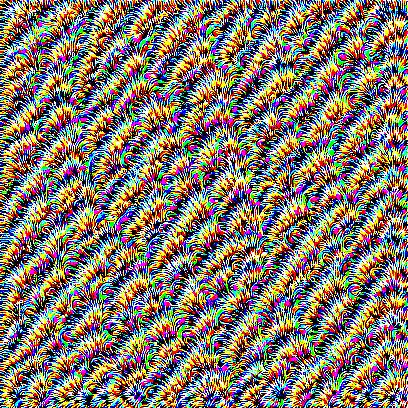}
        \caption*{Synthesized Image, Neuron: 413}
    \end{subfigure}
    \\
\caption{Generated unbounded adversarial images along with synthesized neuron visualizations for the top 20 to 30 attacked neurons. The positions of the neurons are listed below each image. Note that, the top-k neurons were selected based on the transferability to the  DenseNet121~\cite{densenet} as the held-out model after initial lightweight training}
\label{fig:vis3}
\end{figure*}

\begin{figure*}[ht!]
    \centering
    \captionsetup{justification=centering, margin=1cm}
    \begin{subfigure}[b]{0.22\textwidth}
        \centering
        \includegraphics[width=\textwidth]{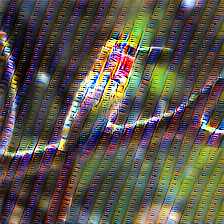}
        \caption*{{ Adversarial Image, Neuron: 409}}
    \end{subfigure}
    \begin{subfigure}[b]{0.22\textwidth}
        \centering
    \includegraphics[width=\textwidth]{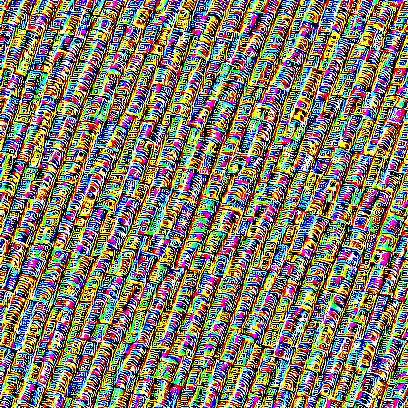}
        \caption*{{ Synthesized Image, Neuron: 409}}
    \end{subfigure}
    \begin{subfigure}[b]{0.22\textwidth}
        \centering
        \includegraphics[width=\textwidth]{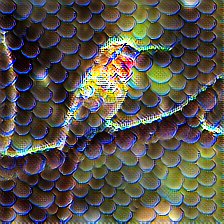}
        \caption*{Synthesized Image, Neuron: 174}
    \end{subfigure}
    \begin{subfigure}[b]{0.22\textwidth}
        \centering
    \includegraphics[width=\textwidth]{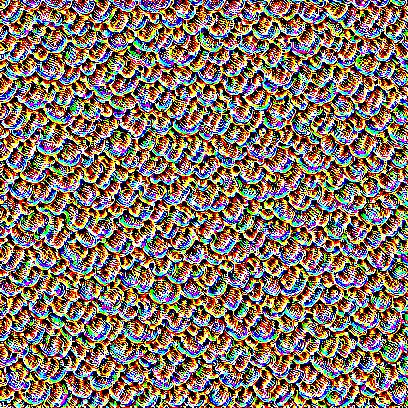}
        \caption*{Synthesized Image, Neuron: 174}
    \end{subfigure}
    \\
    \begin{subfigure}[b]{0.22\textwidth}
        \centering
        \includegraphics[width=\textwidth]{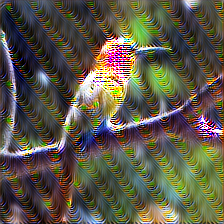}
        \caption*{Adversarial Image, Neuron: 119}
    \end{subfigure}
    \begin{subfigure}[b]{0.22\textwidth}
        \centering
    \includegraphics[width=\textwidth]{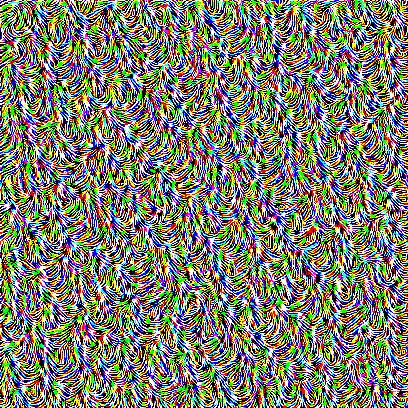}
        \caption*{Synthesized Image, Neuron: 119}
    \end{subfigure}
    \begin{subfigure}[b]{0.22\textwidth}
        \centering
        \includegraphics[width=\textwidth]{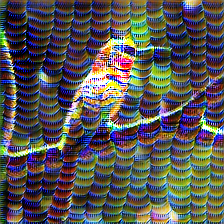}
        \caption*{Synthesized Image, Neuron: 208}
    \end{subfigure}
    \begin{subfigure}[b]{0.22\textwidth}
        \centering
    \includegraphics[width=\textwidth]{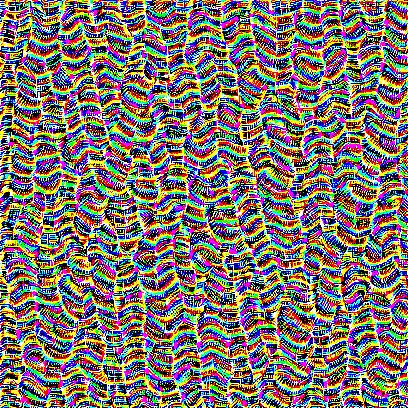}
        \caption*{Synthesized Image, Neuron: 208}
    \end{subfigure}
    \\
     \begin{subfigure}[b]{0.22\textwidth}
        \centering
        \includegraphics[width=\textwidth]{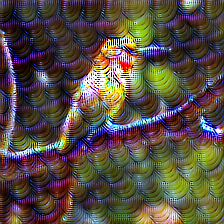}
        \caption*{Adversarial Image, Neuron: 495}
    \end{subfigure}
    \begin{subfigure}[b]{0.22\textwidth}
        \centering
    \includegraphics[width=\textwidth]{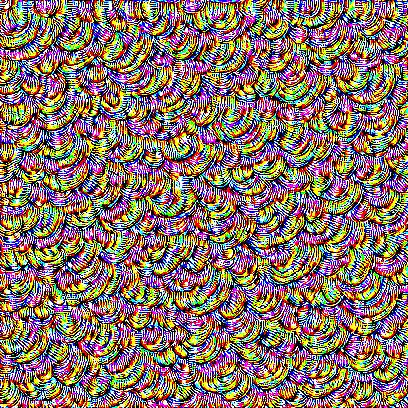}
        \caption*{Synthesized Image, Neuron: 495}
    \end{subfigure}
    \begin{subfigure}[b]{0.22\textwidth}
        \centering
        \includegraphics[width=\textwidth]{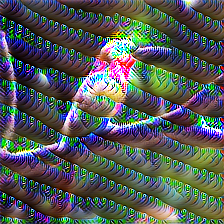}
        \caption*{Synthesized Image, Neuron: 214}
    \end{subfigure}
    \begin{subfigure}[b]{0.22\textwidth}
        \centering
    \includegraphics[width=\textwidth]{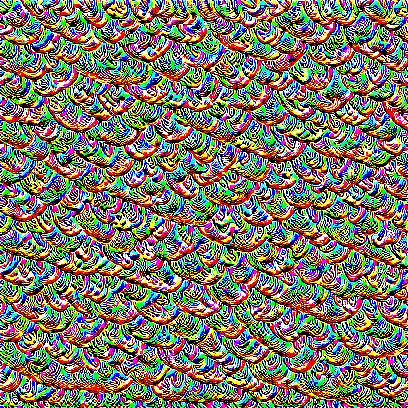}
        \caption*{Synthesized Image, Neuron: 214}
    \end{subfigure}
    \\
    \begin{subfigure}[b]{0.22\textwidth}
        \centering
        \includegraphics[width=\textwidth]{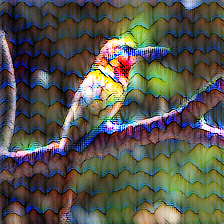}
        \caption*{Adversarial Image, Neuron: 7}
    \end{subfigure}
    \begin{subfigure}[b]{0.22\textwidth}
        \centering
    \includegraphics[width=\textwidth]{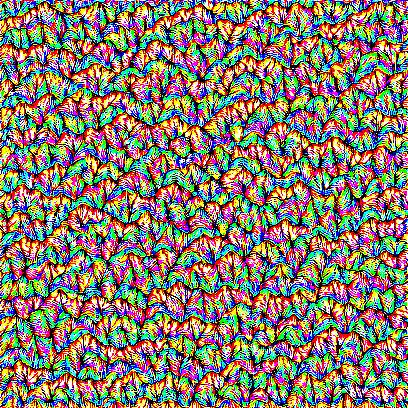}
        \caption*{Synthesized Image, Neuron: 7}
    \end{subfigure}
    \begin{subfigure}[b]{0.22\textwidth}
        \centering
        \includegraphics[width=\textwidth]{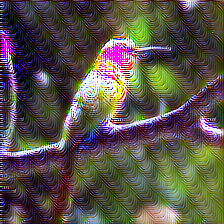}
        \caption*{Synthesized Image, Neuron: 435}
    \end{subfigure}
    \begin{subfigure}[b]{0.22\textwidth}
        \centering
    \includegraphics[width=\textwidth]{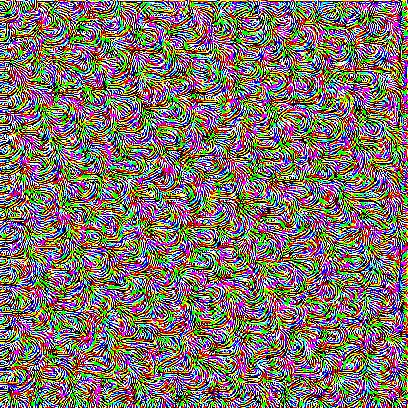}
        \caption*{Synthesized Image, Neuron: 435}
    \end{subfigure}
    \\
    \begin{subfigure}[b]{0.22\textwidth}
        \centering
        \includegraphics[width=\textwidth]{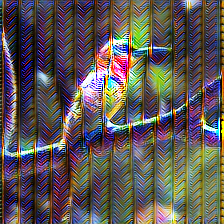}
        \caption*{Adversarial Image, Neuron: 504}
    \end{subfigure}
    \begin{subfigure}[b]{0.22\textwidth}
        \centering
    \includegraphics[width=\textwidth]{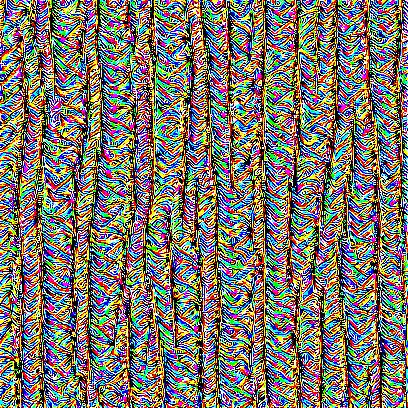}
        \caption*{Synthesized Image, Neuron: 504}
    \end{subfigure}
    \begin{subfigure}[b]{0.22\textwidth}
        \centering
        \includegraphics[width=\textwidth]{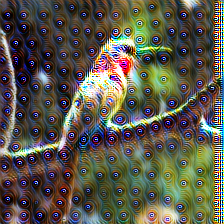}
        \caption*{Synthesized Image, Neuron: 443}
    \end{subfigure}
    \begin{subfigure}[b]{0.22\textwidth}
        \centering
    \includegraphics[width=\textwidth]{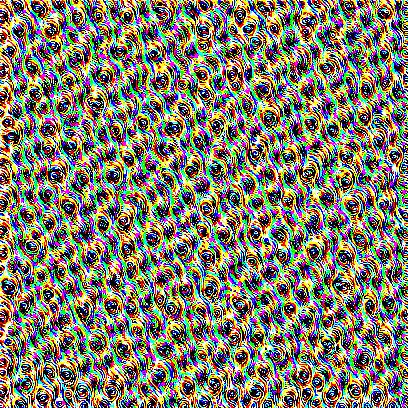}
        \caption*{Synthesized Image, Neuron: 443}
    \end{subfigure}
    \\
\caption{Generated unbounded adversarial images along with synthesized neuron visualizations for the top 30 to 40 neurons. . The positions of the neurons are listed below each image. Note that, the top-k neurons were selected based on the transferability to the  DenseNet121~\cite{densenet} as the held-out model after initial lightweight training}
\label{fig:vis4}
\end{figure*}

We present the visualizations of adversarial images generated by different neuron-specific generators in Figures~\ref{fig:vis1},~\ref{fig:vis2},~\ref{fig:vis3}, and~\ref{fig:vis4}. Next to each adversarial image, we show the synthesized images that is optimized to have large activation magnitudes for the attacked neuron using activation maximization algorithm~\cite{actimax}. A clear visual correlation can be observed between the synthesized images and the adversarial patterns in the generated images, indicating that the perturbations are effectively disrupting neurons associated with specific concepts.

\begin{table*}[t]
\centering
\small
\resizebox{0.6\linewidth}{!}{%
\begin{tabular}{llcccr}
\toprule
Model & Version & Accuracy & Source \\
\midrule
VGG~\cite{vgg} & vgg16 & 71.07\% & TorchVision~\cite{torchvision2016} \\
Resnet~\cite{resnet} & resnet152 & 78.05\% & TorchVision~\cite{torchvision2016} \\
Densenet~\cite{densenet} & densenet121 & 74.72\% & TorchVision~\cite{torchvision2016} \\
Squeezenet1~\cite{squeezenet} & squeezenet1\_1 & 58.58\% & TorchVision~\cite{torchvision2016} \\
Shufflenet~\cite{shufflenet} & shufflenetv2 & 69.52\% & TorchVision~\cite{torchvision2016} \\
Mobilenet~\cite{mobilenet} & mobilenet\_v3 & 74.84\% & TorchVision~\cite{torchvision2016} \\
Mnasnet~\cite{mnasnet} & mnasnet1\_0 & 73.84\% & TorchVision~\cite{torchvision2016} \\
Wide Resnet~\cite{wideresnet} & wide\_resnet50\_2 & 78.74\% & TorchVision~\cite{torchvision2016} \\
Convnext~\cite{convnext} & convnext\_base & 84.05\% & TorchVision~\cite{torchvision2016} \\
Efficientnet~\cite{efficientnet} & efficientnet\_v2\_m & 80.00\% & TorchVision~\cite{torchvision2016} \\
Regnet~\cite{regnet} & regnet\_x\_32gf & 83.62\% & TorchVision~\cite{torchvision2016} \\
Alexnet~\cite{alexnet} & alexnet & 56.88\% & TorchVision~\cite{torchvision2016} \\
ViT~\cite{vit} & vit\_b\_16 & 81.06\% & TorchVision~\cite{torchvision2016} \\
ViT~\cite{vit} & vit\_l\_16 & 80.01\% & TorchVision~\cite{torchvision2016} \\ \midrule
Swin~\cite{swin} & swin\_base\_patch4\_window7\_224 & 85.62\% & timm~\cite{rw2019timm} \\
BeiT~\cite{beit} & beit\_base\_patch16\_224 & 85.68\% & timm~\cite{rw2019timm} \\
DeiT~\cite{deit} & deit3\_base\_patch16\_224 & 84.01\% & timm~\cite{rw2019timm} \\
Mixer~\cite{mixer} & mixer\_b16\_224 & 77.02\% & timm~\cite{rw2019timm} \\
Convmixer~\cite{convmixer} & convmixer\_768\_32 & 80.78\% & timm~\cite{rw2019timm} \\
Efficientvit~\cite{efficientvit} & efficientvit\_m3 & 73.54\% & timm~\cite{rw2019timm} \\
Xception~\cite{xception} & xception & 75.54\% & timm~\cite{rw2019timm} \\
Crossvit~\cite{crossvit} & crossvit\_base\_240 & 82.56\% & timm~\cite{rw2019timm} \\
Csp~\cite{cspresnet} & cspresnet50 & 78.66\% & timm~\cite{rw2019timm} \\
Davit~\cite{davit} & davit & 84.86\% & timm~\cite{rw2019timm} \\
Edgenext~\cite{edgenext} & edgenext & 83.07\% & timm~\cite{rw2019timm} \\
Gcvit~\cite{gcvit} & gcvit\_base & 85.32\% & timm~\cite{rw2019timm} \\
Ghostnet~\cite{ghostnet} & ghostnetv2\_100 & 75.92\% & timm~\cite{rw2019timm} \\
Visformer~\cite{visformer} & visformer\_small & 82.22\% & timm~\cite{rw2019timm} \\
Focalnet~\cite{focalnet} & focalnet\_base\_lrf & 84.54\% & timm~\cite{rw2019timm} \\
Hiera~\cite{hiera} & hiera\_base\_224 & 85.02\% & timm~\cite{rw2019timm} \\
Hrnet~\cite{hrnet} & hrnet\_w32 & 79.03\% & timm~\cite{rw2019timm} \\
Maxvit~\cite{maxvit} & maxvit\_base\_tf\_224 & 85.52\% & timm~\cite{rw2019timm} \\
Convformer~\cite{conformer} & convformer\_s36 & 84.56\% & timm~\cite{rw2019timm} \\
Mobilevit~\cite{mobilevit} & mobilevitv2\_150 & 79.08\% & timm~\cite{rw2019timm} \\
Pnasnet~\cite{pnasnet} & pnasnet5large & 78.52\% & timm~\cite{rw2019timm} \\
Rdnet~\cite{rdnet} & rdnet\_base & 85.14\% & timm~\cite{rw2019timm} \\
Levit~\cite{levit} & levit\_conv\_256 & 82.18\% & timm~\cite{rw2019timm} \\
Mvit~\cite{mvit} & mvitv2\_base & 85.08\% & timm~\cite{rw2019timm} \\
Nf~\cite{nfresnet} & nf\_resnet50 & 79.86\% & timm~\cite{rw2019timm} \\
Coat~\cite{coat} & coat\_lite\_small & 82.76\% & timm~\cite{rw2019timm} \\
Cait~\cite{cait} & cait\_s24\_224 & 84.03\% & timm~\cite{rw2019timm} \\
Dla~\cite{dla} & dla102 & 78.54\% & timm~\cite{rw2019timm} \\
\bottomrule
\end{tabular}%
}\caption{Details of all 41 target models used in our evaluation, along with their clean accuracy on 5K evaluation ImageNet dataset provided by LTP~\cite{ltp}.}
  \label{tab:targetmodels}

\end{table*}

\begin{table*}[t]
\centering
\small
\resizebox{0.5\linewidth}{!}{%
  \begin{tabular}{lllccr}
\toprule
Heldout model & Top-40 Neuron locations  \\\midrule
DenseNet121 &  \begin{tabular}[l]{@{}l@{}} 250,
                94,
                470,
                321,
                329,
                391,
                325,
                259,
                241,
                322,
                312,\\
                367,
                109,
                336,
                27,
                373,
                288,
                228,
                22,
                93,
                40,
                31,
                80,
                191,
                446,
                388,
                148,\\
                124,
                84,
                413,
                409,
                174,
                119,
                208,
                495,
                214,
                7,
                435,
                504,
                443 \end{tabular} \\\midrule
ResNet152 & \begin{tabular}[l]{@{}l@{}} 250,
                94,
                    22,
                    259,
                    93,
                    321,
                    367,
                    325,
                    391,
                    27,
                    470,
                    329,
                    228,
                    119,
                    504,
                    322, \\
                    288,
                    214,
                    307,
                    292,
                    130,
                    40,
                    109,
                    413,
                    312,
                    435,
                    208,
                    336,\\
                    5,
                    495,
                    373,
                    236,
                    446,
                    84,
                    80,
                    124,
                    75,
                    266,
                    241,
                    218 \end{tabular}\\\midrule
VGG16 & \begin{tabular}[l]{@{}l@{}} 329,
        124,
        94,
        475,
        280,
        456,
        373,
        495,
        367,
        68,
        81,
        450,
        208,\\
        388,
        464,
        322,
        41,
        250,
        391,
        130,
        58,
        458,
        321,
        477,
        259,
        337,\\
        325,
        336,
        221,
        218,
        270,
        182,
        470,
        93,
        168,
        80,
        331,
        191,
        446,
        183  \end{tabular} \\
\bottomrule
\end{tabular}}
\caption{{\bf Attacked Neuron Positions.} We provide the list of the top-40 neurons selected based on their transferability performance on the held-out model. Note that all 512 generators were initially trained on just 3.12\% of the dataset, with each generator specifically targeting an individual neuron in layer 18 of the source model VGG16.}
  \label{tab:top40neurons}
\end{table*}

\begin{table*}[h!]
\centering
\resizebox{\textwidth}{!}{%
\begin{tabular}{lcccccccccccccccccccc}
\toprule
Model & 250 & 94 & 470 & 321 & 329 & 391 & 325 & 259 & 241 & 322 & 312 & 367 & 109 & 336 & 27 & 373 & 288 & 228 & 22 & 93 \\
\midrule
ResNet152~\cite{resnet} & 89.10 & \underline{90.66} & 84.34 & 88.26 & 80.44 & \textbf{94.28} & 79.14 & 90.06 & 66.70 & 80.22 & 73.14 & 85.76 & 77.84 & 75.66 & 79.68 & 65.40 & 77.28 & 79.16 & 81.54 & 78.00 \\
DenseNet121~\cite{densenet} & 91.02 & 89.56 & 91.44 & 90.70 & 89.44 & \textbf{94.68} & 82.22 & \underline{91.94} & 85.70 & 84.52 & 88.28 & 86.84 & 84.74 & 89.40 & 81.32 & 75.64 & 85.98 & 83.94 & 80.62 & 77.86 \\
SqueezeNet1~\cite{squeezenet} & 89.46 & 83.04 & 88.24 & 85.22 & 89.56 & 86.94 & 82.46 & 81.16 & 87.04 & 72.98 & \underline{90.44} & 78.38 & 79.50 & \textbf{95.20} & 88.36 & 74.78 & 83.28 & 87.34 & 87.06 & 84.90 \\
ShuffleNet~\cite{shufflenet} & 89.02 & 82.86 & \textbf{91.64} & 80.36 & 84.56 & \underline{91.10} & 83.48 & 90.50 & 75.28 & 80.48 & 86.70 & 84.34 & 70.12 & 84.88 & 87.02 & 65.64 & 84.58 & 84.62 & 82.04 & 87.52 \\
MNasNet~\cite{mnasnet} & 93.32 & 86.32 & \textbf{95.58} & 87.28 & 87.72 & 93.12 & 85.76 & 88.54 & 85.28 & 90.60 & 93.16 & 88.58 & 84.48 & 92.30 & 84.86 & 79.54 & 89.20 & \underline{94.16} & 88.90 & 89.22 \\
PNasNet~\cite{pnasnet} & 63.64 & 36.88 & 58.72 & 52.46 & 56.18 & 57.08 & 43.82 & 52.04 & 30.94 & \underline{66.60} & 43.68 & 45.38 & 52.10 & 65.52 & 55.24 & 36.56 & 62.66 & \textbf{69.38} & 57.30 & 54.72 \\
MobileNet~\cite{mobilenet} & 84.26 & 66.86 & \textbf{86.66} & 71.22 & 80.98 & \underline{86.22} & 75.94 & 76.26 & 57.00 & 71.88 & 71.58 & 72.46 & 54.34 & 80.42 & 82.08 & 55.66 & 78.64 & 79.08 & 79.72 & 79.48 \\
WideResNet~\cite{wideresnet} & 92.20 & 90.86 & 91.82 & 93.14 & 87.60 & \textbf{93.32} & 85.16 & 92.12 & 81.22 & 90.94 & 83.68 & \underline{93.26} & 80.00 & 85.94 & 84.56 & 76.22 & 82.68 & 85.48 & 84.60 & 85.48 \\
RegNet~\cite{regnet} & 91.20 & 94.30 & 93.78 & \textbf{96.78} & 82.66 & \underline{96.10} & 83.62 & 86.86 & 72.66 & 91.88 & 81.74 & 88.72 & 72.50 & 82.98 & 80.08 & 76.28 & 83.04 & 82.72 & 79.32 & 78.28 \\
AlexNet~\cite{alexnet} & 49.82 & 47.28 & 46.68 & 43.86 & 48.40 & 41.98 & 42.94 & 45.28 & 38.04 & 48.92 & 41.12 & 42.20 & 49.22 & \textbf{56.68} & 50.06 & 49.24 & 46.12 & \underline{55.66} & 49.94 & 46.90 \\
NF ResNet~\cite{nfresnet} & 76.86 & 75.34 & 77.98 & 77.94 & 65.46 & \textbf{82.92} & 56.56 & \underline{80.38} & 45.62 & 76.88 & 59.26 & 76.34 & 63.22 & 68.26 & 63.34 & 54.32 & 69.86 & 63.04 & 70.76 & 64.86 \\
DLA & 97.86 & 96.74 & 96.72 & 97.10 & 95.16 & \textbf{98.64} & 94.70 & \underline{98.00} & 92.26 & 96.46 & 94.98 & 97.96 & 89.98 & 91.98 & 95.16 & 94.28 & 93.28 & 93.88 & 93.66 & 94.64 \\
GhostNetV2~\cite{ghostnet} & 82.68 & 75.54 & \textbf{89.84} & 84.04 & 76.50 & \underline{86.80} & 75.06 & 83.50 & 58.90 & 83.82 & 75.10 & 83.08 & 65.74 & 79.78 & 79.52 & 65.00 & 80.42 & 82.24 & 80.54 & 77.70 \\
CSPResNet~\cite{cspresnet} & \underline{83.38} & 78.54 & 83.10 & 78.60 & 67.74 & \textbf{88.98} & 64.08 & 82.80 & 64.46 & 75.36 & 72.68 & 78.08 & 61.24 & 77.24 & 70.92 & 52.32 & 73.28 & 71.20 & 74.10 & 72.32 \\
HRNet~\cite{hrnet} & 97.10 & 95.20 & 94.14 & \textbf{97.46} & 90.32 & \underline{97.34} & 88.60 & 89.08 & 90.66 & 93.78 & 94.30 & 92.36 & 89.48 & 95.48 & 85.06 & 92.10 & 90.82 & 92.52 & 91.22 & 87.44 \\
Xception~\cite{xception} & \underline{73.82} & 62.84 & 66.22 & \textbf{74.48} & 66.10 & 71.66 & 51.88 & 70.46 & 50.14 & 71.38 & 57.42 & 57.56 & 70.32 & 68.82 & 60.60 & 49.60 & 64.40 & 70.86 & 64.68 & 56.52 \\
ViT\_B\_16~\cite{vit} & 17.96 & 14.60 & 14.30 & 14.88 & \underline{18.30} & 16.32 & 14.22 & 15.10 & 11.48 & 16.98 & 13.40 & 13.92 & \textbf{18.38} & 15.48 & 17.32 & 15.30 & 16.40 & 18.06 & 17.20 & 15.56 \\
ViT\_L\_16~\cite{vit} & 18.08 & 14.90 & 17.18 & 16.10 & \textbf{19.14} & 17.88 & 14.22 & 17.82 & 12.40 & 17.58 & 14.76 & 15.30 & 17.78 & 17.96 & 18.14 & 15.36 & 18.24 & \underline{18.74} & 17.28 & 17.06 \\
Swin~\cite{swin} & 28.78 & 24.52 & \underline{30.94} & 28.86 & 27.26 & \textbf{33.76} & 21.34 & 28.20 & 17.62 & 21.58 & 22.16 & 25.46 & 18.32 & 21.66 & 25.46 & 19.32 & 25.80 & 25.52 & 25.02 & 24.54 \\
BeIT~\cite{beit} & \textbf{30.48} & 16.22 & 25.60 & 22.04 & 23.46 & 27.94 & 18.36 & 24.70 & 12.00 & 24.02 & 16.46 & 17.50 & 21.40 & 19.64 & \underline{29.20} & 18.30 & 21.44 & 28.32 & 25.38 & 21.94 \\
DeiT~\cite{deit} & 20.54 & 17.42 & 19.02 & 19.64 & 20.32 & \underline{22.14} & 17.48 & 19.84 & 11.90 & 20.38 & 14.86 & 17.08 & 18.44 & 17.84 & \textbf{24.08} & 16.92 & 17.50 & 18.68 & 18.60 & 15.76 \\
Cait~\cite{cait} & 27.62 & 20.42 & 24.52 & 22.40 & \textbf{31.72} & \underline{29.08} & 18.22 & 24.24 & 13.48 & 22.52 & 18.46 & 20.54 & 27.36 & 20.46 & 25.22 & 18.24 & 23.92 & 22.66 & 22.92 & 22.14 \\
Davit~\cite{davit} & 31.62 & 31.70 & 37.70 & 36.50 & \textbf{41.20} & \underline{40.84} & 24.46 & 34.48 & 23.46 & 31.14 & 27.16 & 37.10 & 19.66 & 19.94 & 24.90 & 21.22 & 32.60 & 27.98 & 27.72 & 29.50 \\
ConvNext~\cite{convnext} & 70.90 & 72.64 & \underline{75.02} & 71.64 & 63.38 & \textbf{83.56} & 73.76 & 75.00 & 47.54 & 74.94 & 51.76 & 66.60 & 73.76 & 57.08 & 63.24 & 69.12 & 49.94 & 60.04 & 55.50 & 54.38 \\
Mixer~\cite{mixer} & 32.38 & 28.32 & 35.92 & 31.82 & \textbf{42.82} & 35.18 & 28.66 & 30.82 & 24.72 & 33.50 & 29.48 & 29.34 & 29.74 & 30.98 & 35.88 & 25.40 & \underline{36.88} & 31.44 & 32.92 & 30.48 \\
ConvMixer~\cite{convmixer} & \underline{76.12} & 52.04 & 75.24 & 54.44 & 72.02 & \textbf{84.60} & 66.78 & 67.82 & 41.76 & 71.50 & 55.02 & 73.14 & 61.04 & 55.74 & 67.46 & 51.90 & 58.08 & 70.04 & 60.02 & 66.34 \\
CrossViT~\cite{crossvit} & 19.18 & 16.50 & 17.92 & 17.44 & 17.70 & \textbf{21.40} & 15.36 & 17.36 & 12.28 & 16.16 & 16.00 & 14.50 & 18.26 & 17.02 & 17.78 & 15.94 & \underline{19.82} & 17.24 & 17.68 & 15.62 \\
Edgenext~\cite{edgenext} & 73.26 & 64.30 & 69.48 & 74.38 & 64.24 & \textbf{80.80} & 58.90 & 70.72 & 44.98 & \underline{74.54} & 49.44 & 68.72 & 58.90 & 52.40 & 65.96 & 54.38 & 60.10 & 62.02 & 55.52 & 59.90 \\
GCViT~\cite{gcvit} & 44.54 & 38.94 & 44.48 & 40.98 & 35.30 & \textbf{49.38} & 30.38 & 39.60 & 24.32 & 34.20 & 31.36 & \underline{45.92} & 22.00 & 24.22 & 27.88 & 22.92 & 37.62 & 33.92 & 30.86 & 36.12 \\
Visformer~\cite{visformer} & 71.10 & 62.96 & 68.78 & 59.32 & \underline{79.26} & \textbf{80.60} & 55.74 & 67.10 & 39.36 & 66.38 & 47.78 & 69.04 & 40.48 & 40.36 & 58.78 & 41.48 & 53.32 & 57.72 & 59.92 & 56.92 \\
FocalNet~\cite{focalnet} & 45.66 & 36.48 & \underline{47.98} & 44.82 & 37.76 & \textbf{54.32} & 35.60 & 38.32 & 31.02 & 40.14 & 35.12 & 39.82 & 27.88 & 30.84 & 37.34 & 25.24 & 42.38 & 38.36 & 33.40 & 36.38 \\
Hiera~\cite{hiera} & 43.16 & 33.72 & \underline{43.56} & 36.40 & 38.92 & \textbf{44.78} & 31.18 & 41.14 & 19.82 & 34.50 & 24.22 & 36.46 & 21.30 & 24.92 & 37.04 & 24.58 & 31.56 & 38.72 & 32.80 & 34.06 \\
MaxVit~\cite{maxvit} & \underline{22.94} & 17.98 & 21.00 & 20.96 & 17.40 & \textbf{25.60} & 17.54 & 19.22 & 15.38 & 17.42 & 19.30 & 19.86 & 16.46 & 15.56 & 16.34 & 15.30 & 20.04 & 18.32 & 17.18 & 18.70 \\
Conformer~\cite{conformer} & 47.72 & 38.44 & 45.98 & 42.24 & \textbf{53.34} & \underline{51.72} & 33.54 & 37.30 & 27.68 & 43.56 & 32.64 & 42.04 & 29.26 & 29.14 & 39.54 & 28.68 & 38.18 & 46.40 & 42.96 & 42.60 \\
MobileViT~\cite{mobilevit} & \underline{92.06} & 91.96 & 88.10 & 91.12 & 68.42 & \textbf{94.42} & 81.82 & 90.68 & 72.34 & 87.40 & 79.12 & 88.36 & 61.04 & 78.04 & 76.06 & 85.48 & 80.62 & 84.84 & 80.42 & 79.52 \\
RDNet~\cite{rdnet} & \underline{50.72} & 42.44 & 44.28 & 47.88 & 36.58 & \textbf{58.78} & 33.00 & 41.88 & 25.68 & 43.76 & 31.04 & 43.74 & 21.82 & 24.98 & 28.28 & 28.96 & 36.62 & 37.52 & 34.72 & 37.34 \\
LeViT~\cite{levit} & 67.52 & 58.20 & 66.52 & 62.00 & 63.98 & \textbf{78.00} & 61.14 & \underline{75.12} & 37.40 & 62.40 & 46.94 & 66.24 & 43.46 & 46.66 & 62.48 & 39.78 & 55.86 & 58.26 & 60.10 & 57.04 \\
MViT~\cite{mvit} & 35.38 & 30.04 & 37.50 & 34.04 & 36.40 & \textbf{43.42} & 24.24 & 36.56 & 19.64 & 30.90 & 27.14 & \underline{38.20} & 20.28 & 23.00 & 25.56 & 21.78 & 32.66 & 28.44 & 29.96 & 31.32 \\
Coat~\cite{coat} & 52.14 & \underline{55.50} & 51.10 & 52.22 & 36.20 & \textbf{70.70} & 42.32 & 45.70 & 32.08 & 39.02 & 34.52 & 46.00 & 30.94 & 30.28 & 34.56 & 30.14 & 45.92 & 38.98 & 34.86 & 39.74 \\
EfficientViT~\cite{efficientvit} & 60.04 & 44.78 & 66.24 & 49.00 & 62.44 & \underline{68.62} & 51.62 & 65.74 & 41.48 & 52.02 & 54.46 & 59.30 & 41.32 & 61.32 & \textbf{70.62} & 43.32 & 58.10 & 57.48 & 62.30 & 57.30 \\
EfficientNet~\cite{efficientnet} & 60.64 & 48.00 & \textbf{64.40} & 49.48 & 55.08 & \underline{63.26} & 43.48 & 55.00 & 36.96 & 51.92 & 44.70 & 47.54 & 35.40 & 54.68 & 47.72 & 36.64 & 50.08 & 54.62 & 54.70 & 48.40 \\
\midrule
Average & 60.62 & 54.29 & 60.24 & 57.06 & 56.38 & 64.35 & 50.95 & 58.01 & 43.38 & 56.22 & 49.87 & 55.93 & 47.30 & 51.73 & 53.77 & 45.08 & 53.88 & 55.36 & 53.56 & 52.79 \\
\bottomrule
\end{tabular}%
}
\caption{{\bf Fooling rate of top-20 generators on the 41 target models.} The attack neuron position of each generator is listed along with the columns. Note that the ranking here is obtained by using DenseNet121~\cite{densenet} as the held-out model after lightweight training. The best performing generator for each target model is highlighted in bold, and the second-best is underlined for each target model. For comparison, baselines LTP~\cite{ltp}, BIA~\cite{bia}, CDA~\cite{cda}, GAP~\cite{gap} achieves average fooling rate of 44.6\%, 42.0\% 40.7\% and 34.7\% over 41 architectures. All the top-20 generators obtained with our method outperform the baselines in all cases. }\label{tab:top20genonallmodels}
\end{table*}

\begin{table*}[h!]
\centering
\resizebox{\textwidth}{!}{%
\begin{tabular}{lcccccccccccccccccccc}
\toprule
Model & 40 & 31 & 80 & 191 & 446 & 388 & 148 & 124 & 84 & 413 & 409 & 174 & 119 & 208 & 495 & 214 & 7 & 435 & 504 & 443 \\
\midrule
ResNet152~\cite{resnet} & 77.40 & 59.46 & 71.60 & 69.08 & 70.58 & 77.58 & 70.36 & 77.84 & 73.18 & 74.28 & 50.26 & 74.76 & \textbf{81.86} & 63.72 & \underline{79.66} & 60.22 & 65.26 & 74.66 & 70.74 & 67.00 \\
DenseNet121~\cite{densenet} & 82.98 & 76.14 & 81.00 & 83.74 & 81.30 & 84.80 & \textbf{86.12} & 84.82 & 78.62 & 76.44 & 62.36 & \underline{84.96} & 82.52 & 70.08 & 83.20 & 67.54 & 74.46 & 77.94 & 73.08 & 80.94 \\
SqueezeNet1~\cite{squeezenet} & 88.50 & 81.90 & \underline{88.58} & 85.72 & 85.04 & 85.54 & \textbf{89.34} & 79.86 & 82.06 & 83.12 & 79.26 & 80.60 & 84.52 & 63.52 & 86.14 & 72.84 & 73.24 & 79.94 & 68.16 & 85.42 \\
ShuffleNet~\cite{shufflenet} & 77.08 & 63.02 & 78.12 & 76.46 & 75.28 & 72.70 & 75.00 & 69.42 & 72.40 & 83.74 & 50.74 & 79.88 & \textbf{87.36} & 51.74 & \underline{84.36} & 69.26 & 64.04 & 83.84 & 67.92 & 70.68 \\
MNasNet~\cite{mnasnet} & 91.56 & 81.46 & 84.56 & 90.60 & 91.68 & \textbf{92.90} & 90.50 & 86.96 & 90.48 & 85.34 & 87.82 & 90.96 & 89.30 & 71.72 & \underline{91.72} & 81.34 & 76.40 & 88.00 & 71.80 & 87.54 \\
PNasNet~\cite{pnasnet} & 54.46 & 31.46 & 53.46 & 38.80 & 37.78 & 55.42 & 33.68 & 48.72 & 56.86 & 54.60 & 35.44 & \underline{63.28} & 62.34 & 43.00 & \textbf{64.08} & 50.70 & 40.84 & 56.48 & 30.78 & 37.92 \\
MobileNet~\cite{mobilenet} & 75.44 & 52.08 & 66.92 & 70.42 & 62.28 & 76.12 & 65.18 & 69.04 & \textbf{81.80} & 77.00 & 63.28 & 62.96 & \underline{78.28} & 58.12 & 75.44 & 63.72 & 64.02 & 75.72 & 50.84 & 66.16 \\
WideResNet~\cite{wideresnet} & \textbf{86.34} & 78.88 & 82.54 & 84.20 & \underline{85.98} & 85.34 & 85.70 & 84.86 & 83.78 & 80.12 & 67.64 & 84.36 & 85.02 & 70.24 & 82.16 & 71.56 & 75.94 & 82.86 & 74.38 & 82.96 \\
RegNet~\cite{regnet} & 82.54 & 75.20 & 79.58 & 82.94 & 80.68 & 83.40 & 83.82 & \textbf{86.92} & 83.06 & 74.70 & 63.60 & 75.72 & 82.68 & 69.62 & \underline{85.48} & 76.34 & 71.98 & 77.58 & 63.52 & 79.02 \\
AlexNet~\cite{alexnet} & 52.66 & 48.66 & 51.64 & 47.66 & 50.34 & 50.14 & 41.00 & 49.56 & \textbf{56.54} & 51.62 & 54.10 & 52.02 & 45.16 & \underline{55.16} & 46.28 & 42.10 & 47.56 & 39.38 & 38.08 & 49.78 \\
NF ResNet~\cite{nfresnet} & 63.94 & 45.60 & 56.58 & 50.26 & 50.12 & 63.12 & 54.76 & 66.86 & \underline{71.12} & 65.46 & 34.94 & \textbf{71.50} & 61.98 & 57.66 & 64.08 & 46.52 & 52.96 & 55.92 & 49.28 & 53.46 \\
DLA & 93.68 & 89.12 & 89.34 & 93.98 & 93.48 & 95.44 & \underline{95.76} & 94.66 & 95.54 & 92.10 & 90.70 & 94.74 & 95.66 & 86.42 & \textbf{96.20} & 90.16 & 87.84 & 94.22 & 84.40 & 92.92 \\
GhostNetV2~\cite{ghostnet} & 80.74 & 51.24 & 67.92 & \textbf{86.42} & 74.28 & 82.60 & 79.88 & 77.52 & 79.28 & 75.90 & 69.52 & 79.06 & 81.02 & 64.12 & \underline{83.26} & 69.44 & 66.20 & 75.02 & 60.60 & 69.36 \\
CSPResNet~\cite{cspresnet} & \textbf{74.20} & 57.64 & 68.72 & 65.66 & 63.90 & 70.02 & \underline{72.34} & 69.26 & 71.24 & 65.36 & 38.90 & 64.00 & 72.14 & 61.56 & 70.34 & 53.18 & 59.28 & 65.50 & 56.26 & 66.30 \\
HRNet~\cite{hrnet} & 94.78 & 92.24 & 92.46 & 93.18 & 91.74 & 95.14 & \textbf{96.28} & \underline{95.74} & 90.54 & 83.20 & 88.68 & 88.74 & 86.74 & 89.82 & 92.76 & 88.22 & 79.24 & 86.14 & 77.78 & 91.36 \\
Xception~\cite{xception} & 65.28 & 46.96 & 54.94 & 55.02 & 54.84 & 61.12 & 55.12 & \underline{66.74} & 65.86 & 57.42 & 49.12 & \textbf{74.86} & 60.72 & 59.42 & 61.84 & 55.32 & 63.30 & 52.18 & 44.64 & 55.66 \\
ViT\_B\_16~\cite{vit} & 14.42 & 13.62 & 15.98 & 14.60 & 14.24 & 18.00 & 13.28 & 15.82 & 18.26 & \underline{18.68} & 16.36 & \textbf{22.56} & 14.56 & 17.48 & 17.04 & 15.34 & 17.30 & 12.64 & 13.14 & 14.56 \\
ViT\_L\_16~\cite{vit} & 15.88 & 15.46 & 17.20 & 14.92 & 15.94 & 18.44 & 13.24 & 17.96 & \underline{20.64} & 19.90 & 15.42 & \textbf{20.96} & 16.74 & 18.68 & 17.88 & 16.64 & 17.12 & 15.06 & 13.72 & 15.08 \\
Swin~\cite{swin} & 20.38 & 20.22 & 21.72 & 17.98 & 16.70 & 23.06 & 18.36 & 22.40 & \underline{26.12} & 22.82 & 14.26 & 20.06 & \textbf{26.58} & 19.58 & 25.76 & 21.64 & 22.78 & 20.50 & 16.08 & 19.22 \\
BeIT~\cite{beit} & 17.14 & 13.94 & 19.42 & 15.12 & 14.50 & 23.46 & 13.96 & 18.24 & \underline{24.10} & \textbf{30.00} & 16.38 & 22.90 & 20.02 & 21.18 & 23.72 & 18.56 & 21.12 & 14.24 & 14.92 & 16.76 \\
DeiT~\cite{deit} & 15.44 & 12.52 & 14.78 & 14.24 & 13.68 & 16.88 & 12.74 & 14.96 & \textbf{19.84} & 17.92 & 12.54 & 17.94 & 17.68 & 16.46 & \underline{18.50} & 15.76 & 16.40 & 14.90 & 14.32 & 13.96 \\
Cait~\cite{cait} & 20.34 & 16.28 & 23.60 & 17.60 & 15.50 & 22.38 & 15.14 & \textbf{27.60} & 26.20 & 24.70 & 15.70 & \underline{26.48} & 21.14 & 22.68 & 24.32 & 18.34 & 20.32 & 17.26 & 14.78 & 16.98 \\
Davit~\cite{davit} & 24.18 & 21.12 & 20.62 & 23.78 & 22.42 & 24.06 & 25.76 & \underline{28.62} & 26.72 & 25.62 & 17.18 & 22.66 & 28.00 & 21.12 & 27.22 & 21.48 & \textbf{28.96} & 25.82 & 18.82 & 23.68 \\
ConvNext~\cite{convnext} & \underline{63.86} & 45.16 & 57.24 & 49.64 & 44.74 & 60.28 & 57.66 & \textbf{84.10} & 58.86 & 55.10 & 43.84 & 59.22 & 58.58 & 57.36 & 59.12 & 54.98 & 58.36 & 48.92 & 52.90 & 59.32 \\
Mixer~\cite{mixer} & 29.56 & 26.30 & 28.46 & 26.54 & 26.02 & 31.26 & 25.26 & 29.34 & 32.88 & \textbf{34.92} & 24.80 & 30.44 & 33.84 & 32.36 & \underline{33.92} & 27.86 & 31.28 & 29.94 & 25.64 & 27.36 \\
ConvMixer~\cite{convmixer} & 64.28 & 45.70 & 52.48 & 50.88 & 56.56 & 55.92 & 48.00 & \textbf{74.74} & 57.68 & 66.12 & 36.38 & \underline{73.74} & 69.60 & 49.58 & 68.88 & 42.38 & 56.20 & 54.64 & 36.88 & 57.82 \\
CrossViT~\cite{crossvit} & 15.88 & 14.30 & 15.70 & 13.94 & 14.78 & 16.26 & 13.04 & 17.32 & \underline{18.46} & 17.60 & 13.50 & \textbf{18.64} & 17.70 & 17.66 & 18.08 & 14.76 & 17.62 & 14.20 & 13.88 & 14.52 \\
Edgenext~\cite{edgenext} & 56.52 & 43.66 & 56.44 & 56.36 & 46.36 & \underline{65.96} & 52.34 & 65.06 & \textbf{67.22} & 57.56 & 42.84 & 61.60 & 56.40 & 55.18 & 65.56 & 50.64 & 52.26 & 48.26 & 45.22 & 50.08 \\
GCViT~\cite{gcvit} & 30.22 & 23.96 & 23.32 & 26.48 & 24.84 & 26.70 & 27.74 & \textbf{39.34} & 34.08 & 26.20 & 18.26 & 26.54 & \underline{39.18} & 26.06 & 34.58 & 24.92 & 30.92 & 29.02 & 19.52 & 28.52 \\
Visformer~\cite{visformer} & 49.90 & 40.92 & 47.00 & 43.70 & 41.90 & 61.82 & 47.02 & 54.00 & \underline{63.68} & 56.44 & 35.62 & 46.24 & \textbf{63.82} & 43.54 & 62.18 & 41.98 & 44.16 & 52.28 & 35.38 & 43.32 \\
FocalNet~\cite{focalnet} & 32.56 & 29.00 & 30.86 & 36.44 & 29.72 & 33.62 & 32.00 & \textbf{44.56} & 35.80 & 30.12 & 23.12 & 29.68 & \underline{44.22} & 25.50 & 40.78 & 28.74 & 32.26 & 33.34 & 22.04 & 36.48 \\
Hiera~\cite{hiera} & 26.24 & 21.48 & 22.20 & 22.76 & 21.76 & \underline{39.68} & 21.48 & 26.50 & 37.98 & 36.04 & 21.32 & 27.62 & 39.58 & 23.90 & \textbf{39.86} & 25.38 & 25.02 & 31.40 & 21.10 & 22.86 \\
MaxVit~\cite{maxvit} & 18.60 & 16.58 & 14.28 & 16.06 & 14.72 & 15.94 & 17.04 & 19.00 & 15.86 & 17.18 & 12.72 & 16.68 & \underline{20.30} & 17.94 & 17.92 & 17.04 & \textbf{20.44} & 16.08 & 13.54 & 16.64 \\
Conformer~\cite{conformer} & 34.94 & 31.72 & 37.32 & 31.64 & 28.08 & \underline{43.10} & 31.36 & \textbf{46.12} & 42.92 & 42.20 & 25.64 & 33.84 & 40.84 & 38.68 & 40.34 & 36.18 & 38.38 & 31.88 & 24.66 & 30.18 \\
MobileViT~\cite{mobilevit} & 77.08 & 69.44 & 76.70 & 82.36 & 78.18 & \underline{85.88} & 80.64 & \textbf{90.80} & 84.96 & 66.94 & 68.46 & 83.98 & 76.54 & 79.54 & 84.96 & 72.88 & 82.84 & 67.14 & 69.42 & 76.64 \\
RDNet~\cite{rdnet} & 33.00 & 24.82 & 27.26 & 27.80 & 26.88 & 33.38 & 27.88 & 34.66 & 33.14 & 29.86 & 21.88 & 27.40 & \textbf{37.90} & 32.58 & \underline{36.16} & 26.62 & 33.52 & 34.90 & 23.30 & 27.58 \\
LeViT~\cite{levit} & 47.30 & 35.56 & 40.02 & 48.36 & 41.72 & 60.04 & 42.58 & 51.94 & \underline{60.72} & 57.92 & 31.44 & 51.76 & 54.94 & 37.72 & \textbf{64.62} & 35.32 & 42.70 & 46.82 & 36.16 & 40.10 \\
MViT~\cite{mvit} & 26.34 & 19.56 & 22.88 & 22.02 & 21.46 & 25.84 & 21.04 & 31.20 & \textbf{35.40} & 25.24 & 16.46 & 26.62 & \underline{31.76} & 27.02 & 28.84 & 21.90 & 28.86 & 26.18 & 19.00 & 21.46 \\
Coat~\cite{coat} & 36.18 & 32.02 & 30.48 & 33.52 & 31.24 & 39.94 & 33.44 & \textbf{43.46} & 42.04 & 36.14 & 26.74 & 36.70 & 41.34 & 37.52 & \underline{42.10} & 30.22 & 33.70 & 32.60 & 29.18 & 32.58 \\
EfficientViT~\cite{efficientvit} & 52.70 & 42.14 & 47.56 & 48.86 & 45.26 & 57.40 & 44.20 & 47.20 & \underline{66.62} & \textbf{67.20} & 40.26 & 48.32 & 63.62 & 44.14 & 61.38 & 42.54 & 41.06 & 55.48 & 38.68 & 51.48 \\
EfficientNet~\cite{efficientnet} & \underline{51.92} & 35.96 & 47.58 & 48.26 & 42.92 & 48.76 & 42.32 & 48.94 & \textbf{60.02} & 45.60 & 27.78 & 45.90 & 46.42 & 41.68 & 47.92 & 40.70 & 37.86 & 41.88 & 31.56 & 41.76 \\
\midrule
Average & 51.62 & 42.50 & 48.27 & 48.24 & 46.33 & 53.16 & 47.62 & 53.72 & 54.70 & 51.42 & 39.64 & 51.73 & 54.11 & 45.39 & 54.84 & 44.42 & 46.68 & 48.31 & 40.15 & 47.21 \\ \bottomrule
\end{tabular}%
}
\caption{{\bf Fooling rate of next 20 generators on the 41 target models.} The attack neuron position of each generator is listed along with the columns. Note that the ranking here is obtained by using DenseNet121~\cite{densenet} as the heldout model after lightweight training.The best performing generator is highlighted in bold and second best is underlined for each target model.For comparison, baselines LTP~\cite{ltp}, BIA~\cite{bia}, CDA~\cite{cda}, GAP~\cite{gap} achieves average fooling rate of 44.6\%, 42.0\% 40.7\% and 34.7\% over 41 architectures.  Among the 20 generators listed here, our method outperform the baselines in 17 out of 20 cases. }\label{tab:top20to40genonallmodels}
\end{table*}

\end{document}